\documentclass[journal]{IEEEtran}
\usepackage{stfloats} 
\usepackage{color}
\usepackage{tabularx}
\usepackage{booktabs}
\usepackage{graphicx}
\usepackage{amsmath}
\usepackage{amssymb}
\usepackage{dsfont}
\usepackage{float} 
\usepackage{algorithm}
\usepackage{algpseudocode}
\usepackage{multirow}
\usepackage{microtype}
\usepackage{hyperref}
\usepackage{stfloats} 
\usepackage{color}
\usepackage{tabularx}
\usepackage{booktabs}
\usepackage{graphicx}
\usepackage{amsmath}
\usepackage{amssymb}
\usepackage{dsfont}
\usepackage{float} 
\usepackage{algorithm}
\usepackage{algpseudocode}
\usepackage{multirow}
\usepackage{microtype}
\usepackage{hyperref}
\usepackage{algpseudocode}    
\usepackage{amsmath}
\usepackage{amssymb}
\usepackage{tabularx}
\usepackage{multirow}

\usepackage[table]{xcolor}
\definecolor{OursBlue}{RGB}{224,242,255}
\usepackage{arydshln}
\usepackage{makecell}
\usepackage{pgffor}
\usepackage{tikz}
\usepackage{arydshln}
\usepackage{amsthm}

\usepackage{graphicx}
\usepackage{subcaption}
\makeatletter
\let\IEEEoriginalmakecaption\@makecaption
\long\def\@makecaption#1#2{%
  \ifx\@captype\@IEEEtablestring
    \footnotesize\bgroup\par\centering
    \@IEEEtabletopskipstrut
    {\normalfont\footnotesize #1.\nobreakspace #2}%
    \par\addvspace{0.5\baselineskip}\egroup
    \@IEEEtablecaptionsepspace
  \else
    \IEEEoriginalmakecaption{#1}{#2}%
  \fi
}
\makeatother

\begin{document}
\raggedbottom
\title{Dataset Distillation Based on Saliency-Driven Prototype Alignment}
\author{Yawen Zou$^{1}$,Wenqi Cai$^{1}$, Guang Li$^{2}$, Ling Xiao$^{2}$, Chunzhi Gu$^{3}$, Chao Zhang$^{1}$ \\
$^{1}$University of Toyama, 
$^{2}$Hokkaido University,
$^{3}$University of Fukui
}
\maketitle

\begin{abstract}
Dataset distillation aims to synthesize compact datasets that can approximate the performance of full-data training while significantly reducing computational and storage costs. However, diffusion-based distillation methods often struggle to preserve structural coherence and generalization, especially in visually complex domains. This issue often stems from latent prototypes that are weakly aligned with class-discriminative regions and contaminated by irrelevant background, thereby degrading generation quality and generalization. To address this limitation, we propose a saliency-driven distillation framework that constructs class-discriminative latent prototypes to enhance representativeness and generalization. The framework proceeds in two stages: (1) ensemble Grad-CAM++ saliency is used to construct prototypes emphasizing class-discriminative regions, and (2) hard-prototype refinement is then applied to construct challenging yet class-consistent prototypes, thereby enhancing discriminability and diversity. Importantly, the diffusion backbones (e.g., LDM and DiT) remain frozen; only lightweight classifiers used for saliency extraction are trained. Extensive experiments across multiple benchmarks demonstrate consistent performance improvements over strong baselines. Code will be released.
\end{abstract} 
\section{Introduction}
\label{sec:intro}

Deep learning continues to advance by scaling model capacity and dataset size, but the resulting demands on computation, storage, and data curation have become critical bottlenecks for both research and deployment. To address these challenges, data-centric approaches such as coreset selection \cite{lee2024coreset,welling2009herding,toneva2018empirical} and dataset distillation \cite{wang2018datasetdistillation,sachdeva2023survey,wang2025edf,liu2025survey} have been widely explored as means to reduce training cost while preserving performance. Coreset selection \cite{yang2024mind,moser2025coreset} selects a compact subset that retains the statistical and structural properties of the original dataset, but it remains fundamentally constrained by the need to sample real data, which limits both condensation efficiency and privacy. In contrast, dataset distillation \cite{wang2018datasetdistillation,wang2022cafe,lei2023survey} aims to synthesize a small surrogate dataset that captures the essential supervisory information of the full dataset, enabling comparable performance with substantially reduced computational overhead.

\begin{figure*}[t] 
\centering 
\includegraphics[width=0.95\textwidth]{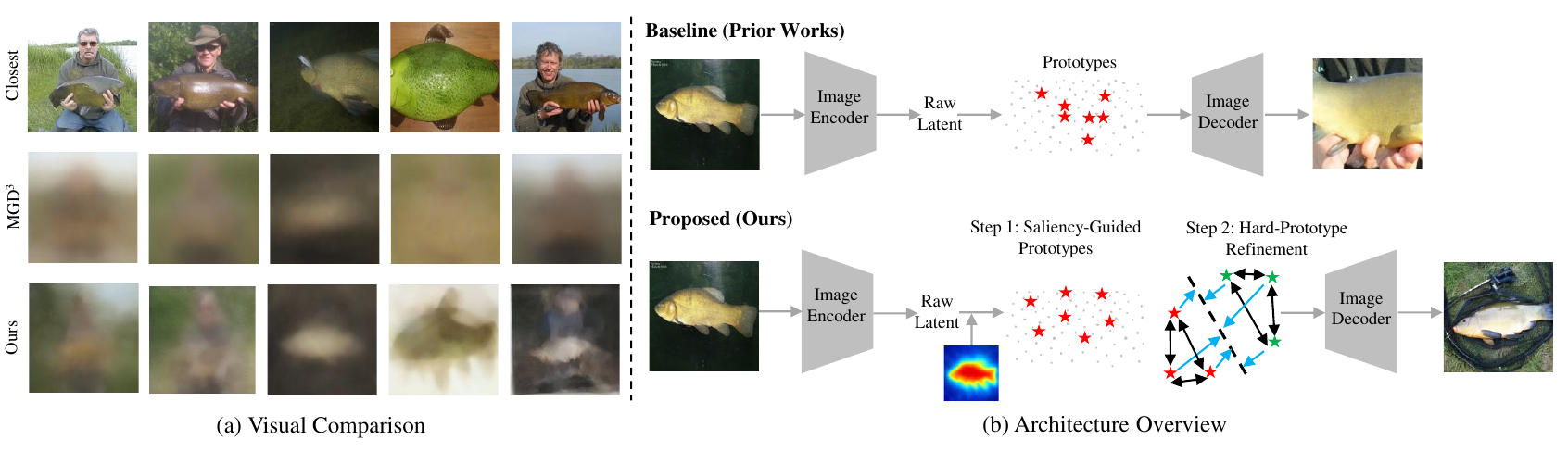} 
\caption{(a) Visual comparison of prototypes. Baseline methods produce background-biased prototypes, while ours focus on class-discriminative regions through saliency guidance. (b) Architecture overview. In contrast to the conventional clustering of raw latents, the proposed framework incorporates saliency masking and hard-prototype refinement to yield more diverse and discriminative prototypes.
} 
\label{fig1} 
\end{figure*}

Conventional dataset distillation approaches can be broadly categorized into three families: meta-learning–based, matching-based, and decoupled frameworks \cite{wang2018datasetdistillation,yu2024teddy,yin2023sre2l,yin2023dataset,ma2025cudd}, which often incur substantial computational costs and generalize poorly across architectures. Generative dataset distillation alleviates these limitations by leveraging GANs \cite{zhao2022synthesizing,wang2023dim,zhao2025hglad} and diffusion models \cite{su2024d4m,gu2024efficient,chan-santiago2025mgd3,zou2025vlcp} to encode dataset knowledge into generators and synthesize compact surrogate datasets in latent space, thereby avoiding the high cost of pixel-space optimization. Among these, diffusion-based methods have shown particular strength due to their ability to generate high-fidelity images. D\textsuperscript{4}M \cite{su2024d4m} and VLCP \cite{zou2025vlcp} are built upon latent diffusion models (LDMs) \cite{rombach2022high}: D\textsuperscript{4}M synthesizes data from class prototypes in the latent space, while VLCP conditions the model on both visual and textual prototypes to improve semantic alignment. In contrast, Minimax \cite{gu2024efficient} and MGD\textsuperscript{3} \cite{chan-santiago2025mgd3} adopt DiT backbones: Minimax fine-tunes a DiT using a minimax objective to produce representative and diverse distilled data, whereas MGD\textsuperscript{3} introduces a mode-guided sampling strategy to generate diverse images without requiring additional fine-tuning on most datasets, except ImageWoof.

While generative distillation has achieved notable success, how latent initializations are optimized to capture class-discriminative features remains largely under-explored. In latent diffusion models (LDMs) \cite{rombach2022high}, latent initialization together with label conditioning jointly influences image fidelity, structural coherence, and diversity. However, existing methods such as D\textsuperscript{4}M, VLCP, and MGD\textsuperscript{3} typically adopt K-Means cluster centers as latent prototypes without considering whether these latents capture category-specific patterns. As illustrated in the architecture overview in Fig.~\ref{fig1} (b), most prior works \cite{su2024d4m,chan-santiago2025mgd3,zou2025vlcp} rely on clustering raw latents produced by a VAE encoder that compresses the entire image without spatial discrimination. This design introduces irrelevant background context into prototype construction. Such a structural limitation weakens alignment between synthesized images and class-relevant features. The visual consequences are evident in Fig.~\ref{fig1} (a). MGD\textsuperscript{3} prototypes may capture coarse, low-frequency shapes with ambiguous object boundaries. Conversely, the closest approach, which selects the nearest sample as the prototype, yields instance-specific representations. This biases the generator toward individual exemplars rather than capturing the broader class structure.

Motivated by this observation, we introduce a saliency-guided and hard-prototype refinement framework designed to prioritize class-discriminative information and capture challenging patterns. Unlike prior methods that cluster raw latents containing background-dominant context, our approach refines latent features via saliency-aware masking before prototype generation.
Specifically, we first extract VAE features and train an ensemble of lightweight classifiers. For each image, Grad-CAM++ \cite{chattopadhay2018grad} maps are aggregated through voting to identify regions critical for classification. The VAE features are then refined by amplifying activations within these regions to emphasize class-discriminative information. We subsequently apply K-Means clustering to the refined latents to obtain class-specific prototypes that better capture discriminative features. Finally, we perform confidence-aware hard-prototype refinement, which adjusts prototype difficulty according to classifier confidence while preserving class-consistent semantics and encouraging intra-class diversity. This strategy yields informative hard prototypes that capture challenging class-consistent patterns and improve downstream performance. The framework is plug-and-play and can be seamlessly integrated with different diffusion backbones, achieving state-of-the-art results on both classification and transfer-learning benchmarks without fine-tuning the diffusion model.

The main contributions are summarized as follows:

\begin{itemize}
\item We introduce a saliency-guided prototype construction strategy that aggregates Grad-CAM++ maps from an ensemble of lightweight classifiers to enhance critical VAE features, enabling prototypes to focus on class-relevant regions in diffusion-based dataset distillation.

\item We further apply hard-prototype refinement to capture challenging patterns, enhancing discriminability and diversity while preserving class-consistent semantics.

\item Extensive experiments across multiple datasets and IPC settings demonstrate that our plug-and-play method consistently improves diffusion-based distillation methods in both classification and transfer learning scenarios.
\end{itemize}

\section{Related work}
\label{sec:Related works}

Dataset distillation aims to synthesize compact surrogate datasets that achieve performance comparable to full-data training at reduced computational cost \cite{wang2018datasetdistillation}. It has found applications in privacy-preserving learning \cite{dong2022privacy,zheng2025dosser,chung2024backdoor}, medical data sharing \cite{kanagavelu2024medsynth,li2024infodist}, and federated learning \cite{goetz2020federated,yan2025fedvck}. Existing approaches can be broadly grouped into non-generative and generative methods.

\begin{figure*}[t] 
\centering 
\includegraphics[width=0.9\textwidth]{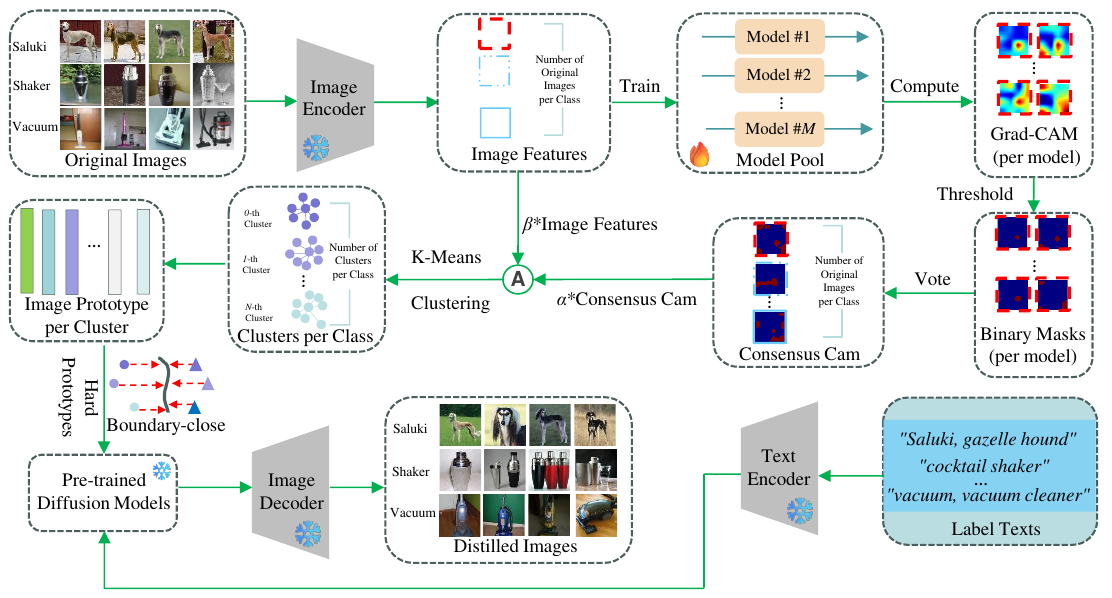} 
\caption{Overview of the proposed saliency-guided dataset distillation framework. The method consists of two stages: (i) constructing class-discriminative latent prototypes using Grad-CAM++ ensemble saliency, and (ii) refining the prototypes according to classifier confidence to construct hard prototypes. \textbf{A} denotes element-wise addition.
} 
\label{fig2} 
\end{figure*}

\noindent\textbf{Non-generative Dataset Distillation Methods.} Early dataset distillation methods adopt a bilevel optimization framework to synthesize data that approximates the training dynamics of full datasets. Wang et al. \cite{wang2018datasetdistillation} used backpropagation through time (BPTT) to unroll the inner-loop updates when optimizing synthetic images, while KIP \cite{nguyen2021kip} replaced the neural learner with kernel ridge regression (KRR) to reduce computational overhead. Subsequent work focused on matching training behavior between real and synthetic data, either by aligning gradients or optimization trajectories \cite{zhao2021datasetcondensation,cazenavette2022dataset,zhong2025mct}, or by minimizing distributional discrepancy through measures such as maximum mean discrepancy (MMD) \cite{zhao2023distribution,zhang2024m3d}.
Decoupled formulations further improved efficiency. SRe\textsuperscript{2}L \cite{yin2023sre2l} decomposed condensation into Squeeze–Recover–Relabel stages, matching batch-normalization statistics and employing soft labels for stability, while CDA \cite{yin2023dataset} introduced curriculum data augmentation to refine gradients from global to local dynamics. More recently, EDF \cite{wang2025edf} enhanced class-relevant regions in pixel space via Grad-CAM++ and continued optimizing synthetic data using gradient matching, yielding gains on more complex datasets.
In contrast, our approach incorporates saliency guidance directly in the latent space of diffusion models, constructing class-discriminative latent prototypes that steer the generative process without performing pixel-level optimization.

\noindent\textbf{Generative Dataset Distillation Methods.} Recent work has incorporated GANs and diffusion models into dataset distillation to synthesize high-quality surrogate datasets with improved scalability and efficiency. Early efforts largely focused on GAN-based strategies. Zhao et al. \cite{zhao2022synthesizing} introduced IT-GAN, which leverages a pretrained GAN and optimizes latent codes of informative examples to preserve learning-relevant information. Wang et al. \cite{wang2023dim} further trained a GAN to align ensemble logits between real and synthetic data, enabling diverse sample synthesis from random noise.
More recently, diffusion-based methods have gained prominence due to their superior image fidelity. Su et al. \cite{su2024d4m} proposed D\textsuperscript{4}M, which inputs K-Means cluster centers as latent prototypes into a latent diffusion model to enhance distillation efficiency and cross-architecture generalization. Zou et al. \cite{zou2025vlcp} integrated category-wise text and image prototypes to guide diffusion-based generation within a vision–language framework. Gu et al. \cite{gu2024efficient} fine-tuned a Diffusion Transformer (DiT) under a minimax objective to generate representative and diverse distilled datasets, while Chan-Santiago et al. \cite{chan-santiago2025mgd3} introduced a mode-guided diffusion strategy that produces diverse images without fine-tuning.
Despite these advances, existing generative approaches typically cluster raw latent without ensuring prototypes capture the intrinsic dataset structure. 

\section{Preliminaries}
\subsection{Dataset Distillation}
Dataset distillation aims to construct a compact synthetic dataset
\( S=\{(\tilde{x}_i,\tilde{y}_i)\}_{i=1}^{N_S} \)
from a large-scale dataset
\( T=\{(x_i,y_i)\}_{i=1}^{N_T} \), where \( N_S \ll N_T \).
The objective is to enable a model trained on \(S\) to achieve performance
comparable to one trained on \(T\). Formally, let \(A(\cdot)\) denote the accuracy
on the test dataset, then $
A(f_S) \approx A(f_T)$,
where \( f_S \) and \( f_T \) are models learned on \(S\) and \(T\), respectively.
The size of the distilled dataset is typically specified in terms of images per class (IPC), i.e., $|S|=\mathrm{IPC}\times C$ for $C$ classes.
After distillation, standard networks are trained on \(S\) and evaluated on the
test set of $T$ to assess how well the synthetic dataset preserves the essential information of the original data.
\subsection{Latent Diffusion Model}
The latent diffusion model (LDM) is a variant of diffusion models in which data generation is performed in the latent space rather than pixel space. A variational autoencoder (VAE) with an encoder-decoder pair ($E$, $D$) is typically used to map images to and from this latent space, i.e., $z=E(x)$ and $\hat{x}=D(z)$, where $z$ is the derived latent. Given a conditioning vector $c$ (e.g., class labels), the training of LDM follows a simplified denoising objective that minimizes the error between the predicted and ground-truth noise $\epsilon$:
\begin{equation}
\mathcal{L}(\theta)=\bigl\|\epsilon-\epsilon_\theta(z_t,c)\bigr\|_2^2,
\end{equation}
where $z_t$ denotes the noisy latent at timestep $t$, and $\epsilon_\theta$ is the noise-prediction network. 

\begin{algorithm}[!t]
\caption{Saliency-Guided Prototype Construction}
\label{alg:sgpc}
\begin{algorithmic}[1]
\State \textbf{Input:} Encoder $E$; classifier ensemble $\mathcal{F}=\{f^{(m)}\}_{m=1}^{M}$; dataset $\mathcal{D}$; CAM operator $\texttt{CAM}(\cdot)$; thresholds $(\tau_{\text{cam}}, \upsilon)$; IPC per class
\State \textbf{Output:} Class-wise prototypes $\mathcal{P}$
\State Initialize per-class latent buffers $\mathcal{Z}_y \gets \varnothing,\;\forall y$
\For{\textbf{each} $(x,y)\in\mathcal{D}$}
    \State $z \gets E(x)$ \Comment{$z\in\mathbb{R}^{4\times H\times W}$}
    \For{$m = 1$ \textbf{to} $M$}
        \State $A^{(m)} \gets \texttt{CAM}(f^{(m)}, z)$
        \State $B^{(m)}(i,j) \gets \mathbf{1}\!\big[A^{(m)}(i,j) \ge \tau_{\text{cam}}\big]$
    \EndFor
    \State $R(i,j) \gets \mathbf{1}\!\big[\sum_{m=1}^{M} B^{(m)}(i,j) \ge \upsilon\big]$ \Comment{voting}
    \State $\tilde{z} \gets \beta z + \alpha (z \odot R)$ \Comment{Enhancing salient regions}
    \State Append $\tilde{z}$ to $\mathcal{Z}_y$
\EndFor
\State \textit{/* Prototype generation via K-Means */}
\State $\mathcal{P} \gets \varnothing$
\ForAll{classes $y$}
    \State $X_y \gets \{\mathrm{vectorize}(\tilde{z}) : \tilde{z}\in\mathcal{Z}_y\}$
    \State $\{c_y^{(1)}, \dots, c_y^{(\mathrm{IPC})}\} \gets \texttt{KMeans}(X_y,\ K{=}\mathrm{IPC})$
    \State $\mathcal{P} \gets \mathcal{P} \cup \{\mathrm{reshape}(c_y^{(j)},shape(z)): j=1,\dots,\mathrm{IPC}\}$
\EndFor
\State \textbf{return} $\mathcal{P}$
\end{algorithmic}
\end{algorithm}

\begin{table*}[t]

\centering
\setlength{\tabcolsep}{10pt}        
{\small                     

\begin{tabular}{l|ccc|ccc}
\toprule
 & \multicolumn{3}{c|}{\textbf{ImageNette}} & \multicolumn{3}{c}{\textbf{ImageIDC}} \\
\cmidrule(lr){2-4}\cmidrule(lr){5-7}
\textbf{Method} & \textbf{IPC = 10} & \textbf{IPC = 20} & \textbf{IPC = 50} & \textbf{IPC = 10} & \textbf{IPC = 20} & \textbf{IPC = 50} \\
\midrule
Random                & 54.2{$\pm$1.6} & 63.5{$\pm$0.5} & 76.1{$\pm$1.1} & 48.1{$\pm$0.8} & 52.5{$\pm$0.9} & 68.1{$\pm$0.7} \\
DiT~\cite{peebles2023scalable}                   & 59.1{$\pm$0.7} & 64.8{$\pm$1.2} & 73.3{$\pm$0.9} & 54.1{$\pm$0.4} & 58.9{$\pm$0.2} & 64.3{$\pm$0.6} \\
DM~\cite{zhao2023distribution}                    & 60.8{$\pm$0.6} & 66.5{$\pm$1.1} & 76.2{$\pm$0.4} & 52.8{$\pm$0.5} & 58.5{$\pm$0.4} & 69.1{$\pm$0.8} \\
MinMax~\cite{gu2024efficient}               & 62.0{$\pm$0.2} & 66.8{$\pm$0.4} & 76.6{$\pm$0.2} & 53.1{$\pm$0.2} & 59.0{$\pm$0.4} & 69.6{$\pm$0.2} \\
DMGD~\cite{wang2026dmgd}               & 66.0{$\pm$0.3} & 72.1{$\pm$0.5} & {78.8{$\pm$0.9}} & {55.0{$\pm$0.7}} & {61.4{$\pm$0.7}} & 71.0{$\pm$1.3} \\
D\textsuperscript{4}M~\cite{su2024d4m} & 61.3{$\pm$1.3} & 71.2{$\pm$2.0} & 77.5{$\pm$0.8} & 54.1{$\pm$1.1} & 62.2{$\pm$0.2} & {72.8{$\pm$1.5}} \\

D\textsuperscript{4}M+Ours & \underline{67.5{$\pm$1.8}} & \underline{72.6{$\pm$1.9}} & 79.0{$\pm$0.1} & \underline {58.1{$\pm$1.9}} & \textbf{64.2{$\pm$0.2}} & \underline{73.6{$\pm$1.5}} \\
MGD\textsuperscript{3}~\cite{chan-santiago2025mgd3}& 66.4{$\pm$2.4} & 71.2{$\pm$0.5} & \underline {79.5{$\pm$1.3}} & 55.9{$\pm$2.1} & 61.9{$\pm$0.9} & 72.1{$\pm$0.8} \\

MGD\textsuperscript{3}+Ours & \textbf{67.9{$\pm$0.8}} & \textbf{73.9{$\pm$2.8}} & \textbf{80.9{$\pm$0.8}} & \textbf{59.7{$\pm$1.5}} & \underline{63.8{$\pm$1.6}} & \textbf{74.1{$\pm$1.5}} \\
\bottomrule
\end{tabular}
}
\par\vspace{6pt}
\caption{Quantitative evaluation on ImageNette and ImageIDC against state-of-the-art methods under different IPC settings (all at 256 ${\times}$ 256 resolution) with ResNetAP-10. The best mean is in bold, and the second best is underlined.}
\label{tab3}
\end{table*}

\section{Method}
We propose a two-stage distillation framework that first constructs prototypes that are class-discriminative and then refines them into hard prototypes, thereby improving both representativeness and discriminability. As illustrated in Fig.~\ref{fig2}, we begin by encoding images using a VAE and training an ensemble of lightweight classifiers on the latent features. The Grad-CAM++ maps from this ensemble are aggregated through majority voting to identify salient, class-relevant regions. We then enhance activations within these regions to obtain saliency-guided latent features, followed by K-Means clustering to construct class-specific prototypes. Finally, we perform hard-prototype refinement by adjusting each prototype toward a moderate-confidence regime through classifier-guided updates, while intra-class repulsion encourages diversity. Together, these stages yield representative and diverse prototypes for downstream distillation.

\subsection{Saliency-Guided Prototype Construction}
\label{sec:4.1}
Unlike prior works \cite{su2024d4m,chan-santiago2025mgd3} that directly cluster raw VAE-extracted latents, we introduce a saliency-guided module that emphasizes class-discriminative information before prototype construction. Specifically, we train an ensemble $\{f^{(m)}\}_{m=1}^{M}$ of M lightweight classifiers (e.g., ResNet-18) using the standard cross-entropy loss:
\begin{equation}
\mathcal{L}^{(m)}=\mathrm{CE} \ \!\big(f^{(m)}(z),\,y\big),
\end{equation}
where the latent $z = E(x) \in \mathbb{R}^{4\times H\times W}$ is obtained from the VAE encoder $E(\cdot)$.
The ensemble encourages saliency stability across random initializations, reducing variance in Grad-CAM++ estimates.

For each image, Grad-CAM++ maps are extracted from the third layer of each classifier, upsampled to $(H, W)$ via bilinear interpolation, and smoothed using a $3 \times 3$ averaging filter to mitigate speckle noise. After max-normalization, each map is binarized, and a spatial position $(i, j)$ is activated if it receives at least $\upsilon$ positive votes among the $M$ classifiers. The resulting aggregated saliency mask $R \in \{0,1\}^{H\times W}$ highlights class-relevant regions. Guided by $R$, we refine the latent representation as:
\begin{equation}
\tilde z \;=\; \beta\, z \;+\; \alpha\, (z \odot R),
\end{equation}
where $\odot$ denotes the Hadamard product, and $(\alpha, \beta)$ control the balance between enhanced class-discriminative activations and original latent features. This refinement amplifies activations within these salient regions, resulting in more informative features for prototype construction.

To identify class prototypes, we perform K-Means clustering on the saliency-refined latent features $\tilde{z}$ for each class, where the number of clusters is set to the target IPC. Due to saliency guidance, the resulting centroids capture discriminative regions while mitigating background bias, resulting in compact and informative latent representations for each class. The complete saliency-guided prototype construction procedure is summarized in Algorithm~\ref{alg:sgpc}.

\subsection{Hard-Prototype Refinement}
\label{sec:4.2}
Although saliency guidance highlights features that distinguish each
class, some prototypes remain too easy and provide little useful
challenge for downstream training. We therefore define hard
prototypes as class-consistent representations with moderate
classifier confidence. We refine any prototype whose confidence lies
outside a preset interval and apply intra-class repulsion to promote
diverse coverage of class-specific patterns.

At iteration $r$, given a prototype $c^r$ with label $y$, we compute
its logits $\ell^r=f(c^r)$ and class confidence
$p^r=\operatorname{softmax}(\ell^r)_y$. Let $\mathcal{R}_y$ denote
the retained class-$y$ prototypes whose confidence lies
within $[p_{\mathrm{low}},p_{\mathrm{high}}]$. We define the
intra-class separation term as
\begin{equation}
\mathcal{D}_{\mathrm{rep}}(c^r)
=
\frac{1}{|\mathcal{R}_y|}
\sum_{c_k\in\mathcal{R}_y}
\bigl(1-\cos(c^r,c_k)\bigr)^2,
\end{equation}
where $\mathcal{D}_{\mathrm{rep}}(c^r)=0$ if
$\mathcal{R}_y$ is empty.

The confidence-dependent refinement objective is
\begin{equation}
\mathcal{J}(c^r;p^r)
=
\begin{cases}
\lambda_{\mathrm{ce}}\operatorname{CE}(\ell^r,y)
+\lambda_{\mathrm{rep}}\mathcal{D}_{\mathrm{rep}}(c^r),
& p^r>p_{\mathrm{high}}, \\[2mm]
-\lambda_{\mathrm{ce}}\operatorname{CE}(\ell^r,y)
+\lambda_{\mathrm{rep}}\mathcal{D}_{\mathrm{rep}}(c^r),
& p^r<p_{\mathrm{low}}.
\end{cases}
\end{equation}

For prototypes whose confidence lies outside the target interval, we
apply the following iterative signed-gradient refinement:
\begin{equation}
c^{r+1}
=
c^r+\eta^r
\operatorname{sign}\!\left(
\nabla_c\mathcal{J}(c^r;p^r)
\right),
\qquad
p^r\notin[p_{\mathrm{low}},p_{\mathrm{high}}],
\end{equation}
where
$\eta^r=\eta\left(1+\frac{r}{T-1}\right)$.
The classification term encourages lower confidence for overly easy
prototypes and higher confidence for insufficiently confident ones,
while the separation term promotes intra-class diversity. Refinement
stops once the confidence enters the target interval, after which the
refined prototype is added to $\mathcal{R}_y$.

\begin{table*}[t]
\centering
{\small
\setlength{\tabcolsep}{1.6pt}

\begin{tabular}{l l c c c c c c c c c c c}
\toprule
\textbf{IPC (Ratio)} & \textbf{Test Model} & \textbf{Random} & \textbf{Herding} & \textbf{DiT} & \textbf{DM} & \textbf{IDC-1} & \textbf{GLaD} & \textbf{MinMax} & \textbf{DMGD} & \textbf{MGD$^3$} & \textbf{Ours} & \textbf{Full} \\
\midrule

\multirow{3}{*}{10 (0.8\%)} &
ConvNet-6
& 24.3{$\pm$}1.1
& 26.7{$\pm$}0.5
& 34.2{$\pm$}1.1
& 26.9{$\pm$}1.2
& 33.3{$\pm$}1.1
& 33.8{$\pm$}0.9
& \textbf{37.0{$\pm$}1.0}
& 32.9{$\pm$}0.5
& 34.7{$\pm$}1.1
& \underline{35.1{$\pm$}1.5}
& 86.4{$\pm$}0.2 \\

& ResNetAP-10
& 29.4{$\pm$}0.8
& 32.0{$\pm$}0.3
& 34.7{$\pm$}0.5
& 30.3{$\pm$}1.2
& 39.1{$\pm$}0.5
& 32.9{$\pm$}0.9
& 39.2{$\pm$}1.3
& 39.2{$\pm$}1.3
& \underline{40.4{$\pm$}1.9}
& \textbf{42.1{$\pm$}2.5}
& 87.5{$\pm$}0.5 \\

& ResNet-18
& 27.7{$\pm$}0.9
& 30.2{$\pm$}1.2
& 34.7{$\pm$}0.4
& 33.4{$\pm$}0.7
& 37.3{$\pm$}0.2
& 31.7{$\pm$}0.8
& 37.6{$\pm$}0.9
& \underline{41.6{$\pm$}0.2}
& 38.5{$\pm$}2.5
& \textbf{43.5{$\pm$}3.5}
& 89.3{$\pm$}1.2 \\

\midrule
\multirow{3}{*}{20 (1.6\%)} &
ConvNet-6
& 29.1{$\pm$}0.7
& 29.5{$\pm$}0.3
& 36.1{$\pm$}0.8
& 29.9{$\pm$}1.0
& 35.5{$\pm$}0.8
& N/A
& 37.6{$\pm$}0.2
& 37.3{$\pm$}0.4
& \underline{39.0{$\pm$}3.5}
& \textbf{40.1{$\pm$}1.8}
& 86.4{$\pm$}0.2 \\

& ResNetAP-10
& 32.7{$\pm$}0.4
& 34.9{$\pm$}0.1
& 41.1{$\pm$}0.8
& 35.2{$\pm$}0.6
& 43.4{$\pm$}0.3
& N/A
& \underline{45.8{$\pm$}0.5}
& \underline{45.8{$\pm$}0.9}
& 43.6{$\pm$}1.6
& \textbf{47.0{$\pm$}1.1}
& 87.5{$\pm$}0.5 \\

& ResNet-18
& 29.7{$\pm$}0.5
& 32.2{$\pm$}0.6
& 40.5{$\pm$}0.5
& 29.8{$\pm$}1.7
& 38.6{$\pm$}0.2
& N/A
& 42.5{$\pm$}0.6
& \underline{47.9{$\pm$}1.0}
& 41.9{$\pm$}2.1
& \textbf{49.4{$\pm$}2.0}
& 89.3{$\pm$}1.2 \\

\midrule
\multirow{3}{*}{50 (3.8\%)} &
ConvNet-6
& 41.3{$\pm$}0.6
& 40.3{$\pm$}0.7
& 46.5{$\pm$}0.8
& 44.4{$\pm$}1.0
& 43.9{$\pm$}1.2
& N/A
& \underline{53.9{$\pm$}0.6}
& 52.3{$\pm$}0.3
& \textbf{54.5{$\pm$}1.6}
& \textbf{54.5{$\pm$}1.0}
& 86.4{$\pm$}0.2 \\

& ResNetAP-10
& 47.2{$\pm$}1.3
& 49.1{$\pm$}0.7
& 49.3{$\pm$}0.2
& 47.1{$\pm$}1.1
& 48.3{$\pm$}1.0
& N/A
& 56.3{$\pm$}1.0
& \underline{60.2{$\pm$}1.1}
& 56.5{$\pm$}1.9
& \textbf{61.5{$\pm$}1.5}
& 87.5{$\pm$}0.5 \\

& ResNet-18
& 47.9{$\pm$}1.8
& 48.3{$\pm$}1.2
& 50.1{$\pm$}0.5
& 46.2{$\pm$}0.6
& 48.3{$\pm$}0.8
& N/A
& 57.1{$\pm$}0.6
& \underline{61.8{$\pm$}2.0}
& 58.3{$\pm$}1.4
& \textbf{63.7{$\pm$}0.8}
& 89.3{$\pm$}1.2 \\

\midrule
\multirow{3}{*}{70 (5.4\%)} &
ConvNet-6
& 46.3{$\pm$}0.6
& 46.2{$\pm$}0.6
& 50.1{$\pm$}1.2
& 47.5{$\pm$}0.8
& 48.9{$\pm$}0.7
& N/A
& \underline{55.7{$\pm$}0.9}
& 55.0{$\pm$}0.8
& 55.1{$\pm$}2.5
& \textbf{57.5{$\pm$}1.4}
& 86.4{$\pm$}0.2 \\

& ResNetAP-10
& 50.8{$\pm$}0.6
& 53.4{$\pm$}1.4
& 54.3{$\pm$}0.9
& 51.7{$\pm$}0.8
& 52.8{$\pm$}1.8
& N/A
& 58.3{$\pm$}0.2
& \underline{63.4{$\pm$}1.0}
& 60.2{$\pm$}2.4
& \textbf{63.6{$\pm$}1.6}
& 87.5{$\pm$}0.5 \\

& ResNet-18
& 52.1{$\pm$}1.0
& 49.7{$\pm$}0.8
& 51.5{$\pm$}1.0
& 51.9{$\pm$}0.8
& 51.1{$\pm$}1.7
& N/A
& 58.8{$\pm$}0.7
& \underline{66.5{$\pm$}1.2}
& 59.7{$\pm$}2.7
& \textbf{67.3{$\pm$}0.3}
& 89.3{$\pm$}1.2 \\

\midrule
\multirow{3}{*}{100 (7.7\%)} &
ConvNet-6
& 52.2{$\pm$}0.4
& 54.4{$\pm$}1.1
& 53.4{$\pm$}0.3
& 55.0{$\pm$}1.3
& 53.2{$\pm$}0.9
& N/A
& \underline{61.1{$\pm$}0.7}
& 59.1{$\pm$}0.7
& 60.1{$\pm$}1.2
& \textbf{61.5{$\pm$}2.2}
& 86.4{$\pm$}0.2 \\

& ResNetAP-10
& 59.4{$\pm$}1.0
& 61.7{$\pm$}0.9
& 58.3{$\pm$}0.8
& 56.4{$\pm$}0.8
& 56.1{$\pm$}0.9
& N/A
& 64.5{$\pm$}0.2
& \underline{67.7{$\pm$}0.6}
& 66.5{$\pm$}1.0
& \textbf{68.5{$\pm$}0.3}
& 87.5{$\pm$}0.5 \\

& ResNet-18
& 61.5{$\pm$}1.3
& 59.3{$\pm$}0.7
& 58.9{$\pm$}1.3
& 60.2{$\pm$}1.0
& 58.3{$\pm$}1.2
& N/A
& 65.7{$\pm$}0.4
& 66.9{$\pm$}0.5
& \underline{68.8{$\pm$}0.7}
& \textbf{70.7{$\pm$}0.7}
& 89.3{$\pm$}1.2 \\

\bottomrule
\end{tabular}
}
\par\vspace{6pt}
\caption{Quantitative evaluation on ImageWoof against state-of-the-art methods under different IPC settings and model architectures (all at 256 $\times$ 256 resolution).
N/A indicates results that are not available in the original papers.}
\label{tab:imagewoof_sota}
\end{table*}

\subsection{Image Synthesis via LDM}
Finally, we employ a latent diffusion model (LDM) conditioned on the refined prototypes and their corresponding class labels to synthesize diverse and representative images. The text encoder $\tau_{\theta}$ embeds the class label $L_c$, providing textual conditioning to the U-Net denoiser during the diffusion process. For a noised prototype latent $z_c^t$ at timestep $t$, the synthesis is formulated as:
\begin{equation}
\textrm{Output} \;=\; D\!\left(U_t\!\left(z_c^t,\, \tau_{\theta}(L_c)\right)\right),
\end{equation}
where $U_t$ denotes the denoiser and $D$ is the VAE decoder. Following prior diffusion-based distillation frameworks such as D\textsuperscript{4}M \cite{su2024d4m} and MGD\textsuperscript{3} \cite{chan-santiago2025mgd3}, we utilize pretrained VAE and diffusion backbones for efficient image generation. In contrast to these methods, our approach conditions the diffusion process on K-Means centroids derived from saliency-refined latents, rather than raw latents, thereby guiding generation toward class-discriminative regions. In practice, all diffusion components remain frozen, and only the lightweight classifiers used for Grad-CAM++ are trained. This design allows the refined prototypes to be seamlessly integrated into existing frameworks such as D\textsuperscript{4}M and MGD\textsuperscript{3}, achieving consistent performance gains with modest additional computational overhead.

\section{Experiments}
\subsection{Datasets}
We evaluate our method on high-resolution benchmarks, including ImageNet-1K and its subsets: ImageIDC, ImageNette, and the more challenging fine-grained ImageWoof dataset. Additionally, results on ImageNet-100 and the low-resolution CIFAR-10 and CIFAR-100 datasets are provided in the Supplementary Material.
\subsection{Implementation Details}

Results are averaged over three runs with different random seeds.
Each lightweight classification model in the ensemble is trained for 50 epochs with a learning rate of $1\times10^{-3}$. We set the ensemble size to $M{=}10$ for all experiments except ImageWoof, where $M{=}5$ is used. We set the confidence interval to $[p_{\text{low}},\, p_{\text{high}}]=[0.4,\, 0.7]$ in all experiments. Additional hyperparameters are provided in Section~2 of the Supplementary Material, and the corresponding sensitivity analysis is provided in Section~3.

\noindent\textbf{Classification Evaluation.}
We consider two evaluation protocols: hard-label and soft-label.
In the hard-label setting, a classifier is trained from scratch on the distilled dataset with hard class labels and evaluated on the original test set with standard augmentations (random resized crop and CutMix).
Following MinMax~\cite{gu2024efficient}, we report results under multiple IPC settings on ImageNet-100, ImageNette, and ImageIDC.
In the soft-label setting, a pre-trained teacher $\phi_T$ produces region-level soft labels \(y_{i,m}=\phi_T(x_{i,m})\) for the \(m\)-th crop of image \(i\).
The student model \(\phi_S\) is then trained on the distilled dataset via cross-entropy against these labels.
Following MGD$^3$~\cite{chan-santiago2025mgd3}, we conduct evaluations on ImageNet-1K.

\noindent\textbf{Transfer Learning Evaluation.} Following KRR-ST \cite{lee2024self}, we first pre-train a randomly initialized feature extractor \(f_{\omega}\) with a linear head \(h_W\) by minimizing the mean-squared error on the distilled dataset \((X_s,Y_s)\):
\[
\min_{\omega,W}\ \tfrac{1}{2}\,\|f_{\omega}(X_s)W - Y_s\|_F^{2}.
\]
We then discard \(h_W\) and fine-tune \(f_{\omega}\) on the target dataset with a newly initialized task-specific head \(h_Q\) using cross-entropy loss.

\subsection{Comparison with SOTA Methods}

We compare our method with diffusion-based approaches, including GLaD~\cite{cazenavette2023generalizing}, Minimax Diffusion~\cite{gu2024efficient}, D$^4$M~\cite{su2024d4m}, MGD$^3$~\cite{chan-santiago2025mgd3}, and DMGD~\cite{wang2026dmgd}, as well as other techniques such as SRe\textsuperscript{2}L~\cite{yin2023sre2l}, IDC-1 \cite{kim2022dataset}, and Herding \cite{welling2009herding}. 

\begin{table*}[t]

\centering
{\small
\setlength{\tabcolsep}{8pt}
\begin{tabular}{lcccccc}
\toprule
Method & CIFAR10 & CIFAR100 & Aircraft & Cars & Dogs & Flowers \\
\midrule
w/o pre~\cite{lee2024self}  & 88.66{$\pm$0.09} & 66.62{$\pm$0.32} & 42.45{$\pm$0.46} & 23.62{$\pm$0.70} & 24.59{$\pm$0.46} & 59.39{$\pm$0.29} \\
Random  & 88.46{$\pm$0.09} & 65.97{$\pm$0.08} & 40.09{$\pm$0.46} & 20.92{$\pm$0.42} &  23.08{$\pm$0.40} & 56.81{$\pm$0.40} \\
FRePo~\cite{zhou2022dataset}   & 87.88{$\pm$0.20} & 65.23{$\pm$0.47} & 39.03{$\pm$0.35} & 20.00{$\pm$0.73}  & 22.05{$\pm$0.45} & 52.50{$\pm$0.51} \\
KRR-ST~\cite{lee2024self}   & \underline{89.33{$\pm$0.19}} & \underline{68.04{$\pm$0.22}} & \underline{57.17{$\pm$0.16}} & \underline{46.95{$\pm$0.37}}  & \underline{35.51{$\pm$0.45}} & \underline{70.45{$\pm$0.34}} \\
\midrule
Ours      & \textbf{90.40{$\pm$0.15}} & \textbf{69.21{$\pm$0.06}} & \textbf{58.68{$\pm$0.75}} & \textbf{47.58{$\pm$0.58}} & \textbf{35.71{$\pm$0.56}} & \textbf{75.94{$\pm$0.39}} \\
\bottomrule
\end{tabular}
}
\par\vspace{6pt}
\caption{Quantitative evaluation on transfer learning from ImageNet-1K distilled data at IPC = 10 (0.8\%). A randomly initialized ConvNet-4 is pre-trained on the distilled dataset and fine-tuned on target datasets. Results are reported as the mean and standard deviation over three runs. The best result is shown in bold, and the second best is underlined. MGD\textsuperscript{3} is our base model.}
\label{tab:transfer_imagenet}
\end{table*}

\begin{table}[t]
\centering

{\small
\setlength{\tabcolsep}{8pt}
\begin{tabular}{lccc}
\toprule
Method & ResNet-18 & ResNet-50 & ResNet-101 \\
\midrule
SRe\textsuperscript{2}L
& 46.8{$\pm$}0.2
& 55.6{$\pm$}0.3
& 60.8{$\pm$}0.5 \\

G-VBSM
& 51.8{$\pm$}0.4
& 58.7{$\pm$}0.3
& 61.0{$\pm$}0.4 \\

EDC
& 58.0{$\pm$}0.2
& 64.3{$\pm$}0.2
& 64.9{$\pm$}0.2 \\

D$^{4}$M
& 55.2{$\pm$}0.1
& 62.4{$\pm$}0.1
& 63.4{$\pm$}0.1 \\

MGD\textsuperscript{3}
& \underline{60.2{$\pm$}0.1}
& \underline{64.6{$\pm$}0.4}
& \textbf{67.7{$\pm$}0.4} \\

Ours
& \textbf{61.9{$\pm$}0.1}
& \textbf{67.0{$\pm$}0.2}
& \underline{67.6{$\pm$}0.2} \\

\midrule
Full
& 69.8
& 80.9
& 81.9 \\
\bottomrule
\end{tabular}
}
\par\vspace{6pt}
\caption{Quantitative comparison of Top-1 accuracy on ImageNet-1K at IPC = 50 using MGD$^{3}$ as the base model.}
\label{tab:imagenet_ipc50}
\end{table}
\noindent\textbf{ImageNette and ImageIDC.} 
To evaluate the effectiveness of our method, we conduct experiments on ImageNette and ImageIDC under multiple IPC settings by integrating our saliency-guided prototype module into D$^{4}$M and MGD$^{3}$ (Tab.~\ref{tab3}).
Our module boosts MGD\textsuperscript{3} with gains of up to +2.7\% on ImageNette and +3.8\% on ImageIDC. When incorporated into D$^{4}$M, it yields maximum gains of 6.2\% and 4.0\% on ImageNette and ImageIDC, respectively. These results demonstrate the effectiveness and broad
applicability of our approach across different datasets and IPC
settings.

\noindent\textbf{Performance on ImageWoof.} We evaluate our method on ImageWoof under five IPC settings using ConvNet-6, ResNetAP-10, and ResNet-18. As shown in Tab.~\ref{tab:imagewoof_sota}, our method achieves the best result among all distilled-data methods in 14 of the 15 settings; the sole exception is ConvNet-6 at IPC = 10, where MinMax performs best, and our method ranks second. Compared with MGD$^3$, our method achieves higher accuracy in 14 of the 15 settings and yields an average improvement of 3.2\%. The gains are particularly pronounced on ResNet-18, where our method surpasses MGD$^3$ by 5.0\%, 7.5\%, 5.4\%, 7.6\%, and 1.9\% at IPC = 10, 20, 50, 70, and 100, respectively. Consistent improvements over MGD$^3$ are also observed on ResNetAP-10 across all IPC settings. Moreover, compared with the state-of-the-art DMGD, our method achieves higher accuracy across all 15 settings, with the largest improvement of 3.8\% on ResNet-18 at IPC = 100. These results demonstrate the effectiveness and strong cross-architecture generalization of our distilled data.

\begin{figure}[t]
\centering
\includegraphics[width=0.8\columnwidth]{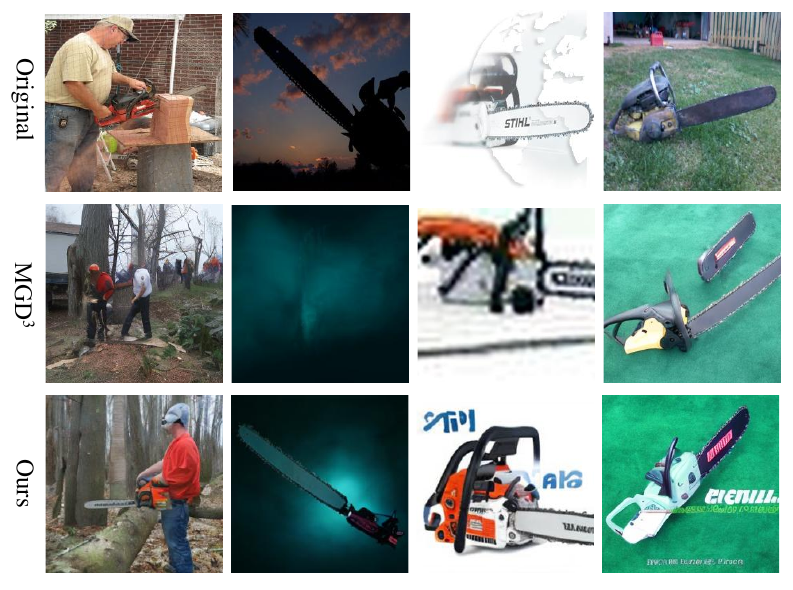}
\caption{Qualitative results on ImageNet-1K. Visualization of synthesized chainsaw samples.}
\label{fig:imagenet1k_qualitative}
\end{figure}

\noindent\textbf{Performance on Larger Models.}
To evaluate scalability with larger models, we use distilled ImageNet-1K data at IPC = 50 and test on ResNet-50 and ResNet-101 in addition to ResNet-18 (Tab. \ref{tab:imagenet_ipc50}).
Compared with the strong diffusion baseline MGD\textsuperscript{3}, our method improves top-1 accuracy by +1.7\% on ResNet-18 and +2.4\% on ResNet-50, and remains comparable on ResNet-101 (-0.1\%).
This gap falls within MGD\textsuperscript{3}'s own standard deviation (±0.4), suggesting comparable performance at this backbone. This may be partly attributable to a capacity-saturation effect at ResNet-101, where the backbone's substantial representational capacity narrows the headroom for further gains from synthesis-quality improvements at this IPC level. This quantitative advantage is supported by the synthesized samples in Fig. \ref{fig:imagenet1k_qualitative}. As exemplified by the chainsaw class, our method produces structurally coherent and detailed images that exhibit superior semantic alignment and visual fidelity relative to the baseline.

\begin{table}[t]

\centering

\begin{tabular}{lccccc}
\toprule
& \multirow{2}{*}{SG}
& \multirow{2}{*}{HP}
& \multicolumn{3}{c}{IPC} \\
\cmidrule(lr){4-6}
& & & 10 & 20 & 50 \\
\midrule
MGD\textsuperscript{3} & -- & -- & 66.4{$\pm$}2.4 & 71.2{$\pm$}0.5 & 79.5{$\pm$}1.3 \\
\midrule
\multirow{3}{*}{Ours} 
& \checkmark & -- & 66.6{$\pm$}0.6 & 73.2{$\pm$}2.5 & 80.6{$\pm$}1.0 \\
& -- & \checkmark & 66.3{$\pm$}1.1 & 71.4{$\pm$}2.5 & 80.4{$\pm$}0.2 \\
& \checkmark & \checkmark & \textbf{67.9{$\pm$}0.8} & \textbf{73.9{$\pm$}2.8} & \textbf{80.9{$\pm$}0.8} \\
\bottomrule
\end{tabular}
\par\vspace{6pt}
\caption{Component ablation on saliency guidance (SG) and hard-prototype (HP) under different IPCs.}

\label{tab:ablation}
\end{table}

\noindent\textbf{Performance on Transfer Learning.}
Following KRR-ST~\cite{lee2024self}, we assess transferability using distilled ImageNet-1K data at IPC = 10. 
As shown in Tab.~\ref{tab:transfer_imagenet}, our method achieves the best accuracy on six  downstream datasets, with an average gain of +1.7\%  and the largest improvement on Flowers (+5.5\%). 
These results demonstrate the generalization and transferability of our distilled data across both coarse- and fine-grained classification tasks.

\noindent\textbf{Computational Cost.}
Our method keeps the diffusion backbone frozen and only trains lightweight classifiers on VAE latents. On ImageNette, training an ensemble of (M=10) ResNet-18 classifiers takes approximately 30 minutes. Under this configuration, the average generation time at ($256 \times 256$) resolution on a single NVIDIA RTX 5090 GPU is 4.44 s/image, compared with 4.08 s/image for MinMax and 0.60 s/image for MGD\textsuperscript{3}. For ImageWoof, we use a smaller ensemble of (M=5), reducing the total runtime to approximately 22 minutes. In comparison, MGD\textsuperscript{3} requires approximately 35 minutes on ImageWoof, including 30 minutes of DiT fine-tuning. Further runtime details are provided in Tab.~7 of the Supplementary Material.

\section{Ablation Study and Analysis}
\noindent\textbf{Component Ablation.}
To assess the contribution of saliency-guidance (SG) and hard-prototype (HP) refinement, we perform an ablation study, with results summarized in Tab. \ref{tab:ablation}. While SG consistently yields improvements, HP refinement shows a strong synergy with SG. Their combination collectively achieves the best performance, particularly as IPC increases. Our full framework reaches peak accuracies of 67.9\%, 73.9\%, and 80.9\% at IPC = 10, 20, and 50, respectively.
These results indicate that the saliency-guidance and the hard-prototype refinement both contribute as intended to producing informative and diverse distilled datasets.

\begin{table}[t]
{\small
\setlength{\tabcolsep}{2.5pt}
\centering
\begin{tabular}{lcccc}
\toprule
Method & Prec. (\%) $\uparrow$ & Dens. $\uparrow$ & Cov. (\%) $\uparrow$ & FID $\downarrow$ \\
\midrule
MGD$^{3}$        & \textbf{90.73{$\pm$0.76}} &  1.17{$\pm$0.10} & 17.90{$\pm$1.30} & 52.37{$\pm$0.89} \\
MGD$^{3}$ + Ours & 90.13{$\pm$1.30} &  \textbf{1.21{$\pm$0.06}} & \textbf{18.51{$\pm$1.17}} & \textbf{51.57{$\pm$1.34}} \\
D$^{4}$M         & 90.33{$\pm$1.33} &  1.03{$\pm$0.05} & 12.93{$\pm$0.57} & 59.43{$\pm$1.18} \\
D$^{4}$M + Ours  & 90.47{$\pm$1.79} &  1.07{$\pm$0.03} & 13.92{$\pm$0.24} & 56.14{$\pm$1.84} \\
\bottomrule
\end{tabular}
}
\par\vspace{6pt}
\caption{Generation quality on ImageNette at IPC = 50.
Prec., Dens., and Cov. denote Precision, Density, and Coverage, respectively.} 
\label{tab:mgd_d4m_ours_metrics}
\end{table}

\noindent\textbf{Quantitative Evaluation of Synthetic Data.}
We further assess generation quality in
Tab.~\ref{tab:mgd_d4m_ours_metrics} using Precision, Density,
Coverage, and FID, following
\cite{gu2024efficient,ferjad2020icml}. Our method consistently
improves Density and Coverage while reducing FID for both
MGD$^{3}$ and D$^{4}$M. Although Precision slightly decreases
from $90.73\%$ to $90.13\%$ on MGD$^{3}$, it remains comparable
to the baseline. On D$^{4}$M, Precision increases from $90.33\%$
to $90.47\%$. Overall, the improvements in Density, Coverage,
and FID indicate better diversity and distributional coverage
while maintaining comparable sample fidelity.

\section{Conclusion}
In this work, we propose a saliency-guided dataset distillation
framework that constructs class-discriminative latent prototypes
and incorporates confidence-aware hard-prototype refinement to
improve generalization and robustness. Specifically, Grad-CAM++ maps
are aggregated from an ensemble of lightweight classifiers to refine VAE latent features, thereby emphasizing
class-relevant regions.
Subsequently, the hard-prototype refinement stage adjusts prototype
difficulty according to classifier confidence, producing
challenging yet class-consistent prototypes with improved diversity
and discriminability. Extensive experiments demonstrate that the
proposed plug-and-play module improves both
D\textsuperscript{4}M and MGD\textsuperscript{3} across a wide
range of classification and transfer-learning settings without
fine-tuning diffusion backbones.

\noindent\textbf{Limitations and Future Work.}
This study primarily focuses on image classification with preliminary validation on transfer learning tasks. In the future, we plan to adapt the framework to more complex visual tasks such as object detection, segmentation, and video analysis.


\bibliographystyle{unsrt}
\bibliography{reference}

\clearpage

\twocolumn[
\begin{center}
    {\LARGE\bfseries SUPPLEMENTARY MATERIAL\par}
    \vspace{0.8em}

    \vspace{1.2em}
\end{center}
]

This supplementary material accompanies the main manuscript ``Dataset Distillation Based on Saliency-Driven Prototype Alignment''. Section~\ref{sec1} presents the implementation details and
hyperparameter settings for the lightweight classifier ensemble and saliency-guided
module. Section~\ref{sec2} analyzes the saliency-guidance parameters, prototype
refinement thresholds, the joint effects of saliency-guidance and hard-prototype
refinement, and the impact of classifier ensemble size on ImageWoof.
Section~\ref{sec3} reports additional quantitative evaluations on ImageNet-100,
CIFAR-10, and CIFAR-100, together with computational-cost analysis and qualitative
comparisons of synthesized samples and saliency maps. Finally,
Sections~\ref{sec4} and~\ref{sec5} provide t-SNE visualizations and uncurated
synthesized images on ImageNette and ImageIDC, respectively, illustrating the
distributional alignment, visual fidelity, and diversity of the distilled datasets.

\setcounter{section}{0}
\section{Experimental Settings}
\label{sec1}
\subsection{Parameter Settings for Lightweight Classifier Ensemble.}
To obtain stable and class-discriminative Grad-CAM++ maps, we train an ensemble of lightweight classifiers directly on VAE-extracted latent features. Specifically, for ImageNet-1K and its subsets (e.g., ImageNette and ImageNet-100), each input image is resized to $256 \times 256$, resulting in a latent representation $z \in \mathbb{R}^{4 \times 32 \times 32}$ after VAE encoding. In contrast, for low-resolution datasets such as CIFAR-10 and CIFAR-100, we resize the images to $512 \times 512$ before encoding to preserve fine-grained details and ensure sufficient spatial resolution in the latent space ($z \in \mathbb{R}^{4 \times 64 \times 64}$) for precise saliency localization. Each classifier takes \(z\) as input and is optimized using the standard cross-entropy loss. 

To avoid redundant computation, we first cache the latent features of the original datasets to disk before classifier training. This step ensures that the subsequent ensemble training focuses solely on classifier optimization rather than repeated VAE encoding. Subsequently, we train an ensemble of $M$ randomly initialized classifiers to mitigate variance in the Grad-CAM++ maps. For ImageNette, ImageIDC, ImageNet-100, CIFAR-10, and CIFAR-100, we adopt a ResNet-18 backbone, whereas for ImageNet-1K we use a ResNet-AP backbone to better accommodate the increased complexity of the dataset. The full set of training hyperparameters is summarized in Tab.~\ref{tab1}.

\subsection{Parameter Settings for Saliency-Guided Module.}

We next summarize the hyperparameters used in the saliency-guided refinement module introduced in Sec.~4.1 of the main paper, where

\[
\tilde z = \beta z + \alpha (z \odot R)
\]

Here, $z$ denotes the original VAE latent representation, $R$ is the binary saliency mask, and $\alpha$ and $\beta$ control the contributions of the saliency-enhanced and original latent features, respectively. The Grad-CAM++ threshold $\tau_{\mathrm{cam}}$ converts the saliency map produced by each classifier into a binary mask, while the voting parameter $\upsilon$ specifies the minimum number of classifiers required to retain a spatial location in the final mask $R$.

We fix $\beta=1$ in all experiments. We use an ensemble of $M=10$ lightweight classifiers for all datasets except ImageWoof, for which $M=5$ is adopted based on the ensemble-size analysis in Sec.~\ref{sec2}. The remaining parameters, $\tau_{\mathrm{cam}}$, $\alpha$, and $\upsilon$, are selected according to the dataset and base distillation method, as summarized in Tab.~\ref{tab:supp_salieny_params}. For ImageNette, the same configuration is used for both MGD$^{3}$ and D$^{4}$M, whereas different configurations are adopted for the two base methods on ImageIDC. For each dataset--method combination, the parameter configuration is kept fixed across all IPC settings and evaluation architectures.

\begin{table}[t]
\centering

\begin{tabular}{lc}
\toprule
\textbf{Settings}          & \textbf{Value}         \\ \hline
Training batch size        & 128                      \\
Number of training epochs  & 50                      \\
Learning rate              & $1 \times 10^{-3}$  \\
Number of classifiers      & 10 (5 for ImageWoof) \\
Optimizer                  & Adam \\
\bottomrule
\end{tabular}%
\vspace{4pt}
\caption{Parameter settings for lightweight classifier ensemble.}%
\label{tab1}
\end{table}

\begin{table}[t]
\centering
\begin{tabular}{llccc}
\toprule
\textbf{Dataset} & \textbf{Base Method}
& $\boldsymbol{\tau_{\mathrm{cam}}}$
& $\boldsymbol{\alpha}$
& $\boldsymbol{\upsilon}$ \\
\midrule
\multirow{2}{*}{ImageIDC}
& MGD$^{3}$ & 0.6 & 0.3 & 2 \\
& D$^{4}$M   & 0.4 & 0.2 & 2 \\
\midrule
\multirow{2}{*}{ImageNette}
& MGD$^{3}$ & 0.6 & 0.2 & 1 \\
& D$^{4}$M   & 0.6 & 0.2 & 1 \\
\midrule
ImageNet-100 & MGD$^{3}$ & 0.7 & 0.2 & 3 \\
ImageWoof    & MGD$^{3}$ & 0.6 & 0.2 & 1 \\
ImageNet-1K  & MGD$^{3}$ & 0.8 & 0.2 & 1 \\
CIFAR-100    & D$^{4}$M  & 0.4 & 0.3 & 2 \\
CIFAR-10     & D$^{4}$M  & 0.6 & 0.2 & 3 \\
\bottomrule
\end{tabular}
\vspace{4pt}
\caption{Parameter settings for the saliency-guided module under different datasets and base methods. We use $M=10$ lightweight classifiers for all datasets except ImageWoof, for which $M=5$ is adopted.}
\label{tab:supp_salieny_params}
\end{table}

\begin{figure*}[t]
\centering

\begin{subfigure}{0.3\linewidth}
    \centering
    \includegraphics[width=\linewidth]{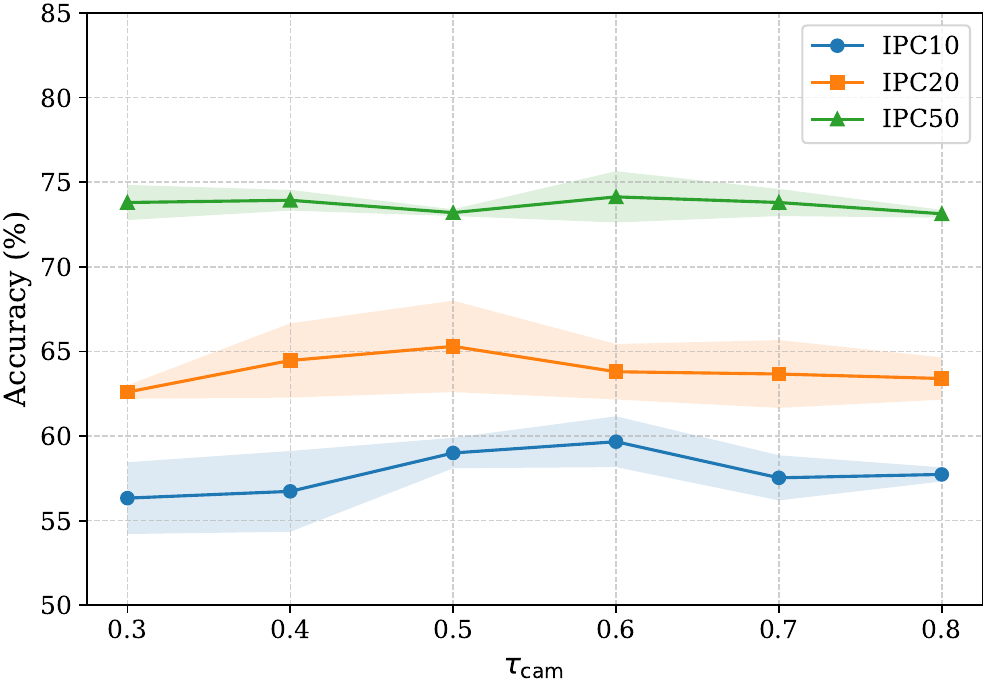}
    \caption{}
    \label{fig:5a}
\end{subfigure}
\hfill
\begin{subfigure}{0.3\linewidth}
    \centering
    \includegraphics[width=\linewidth]{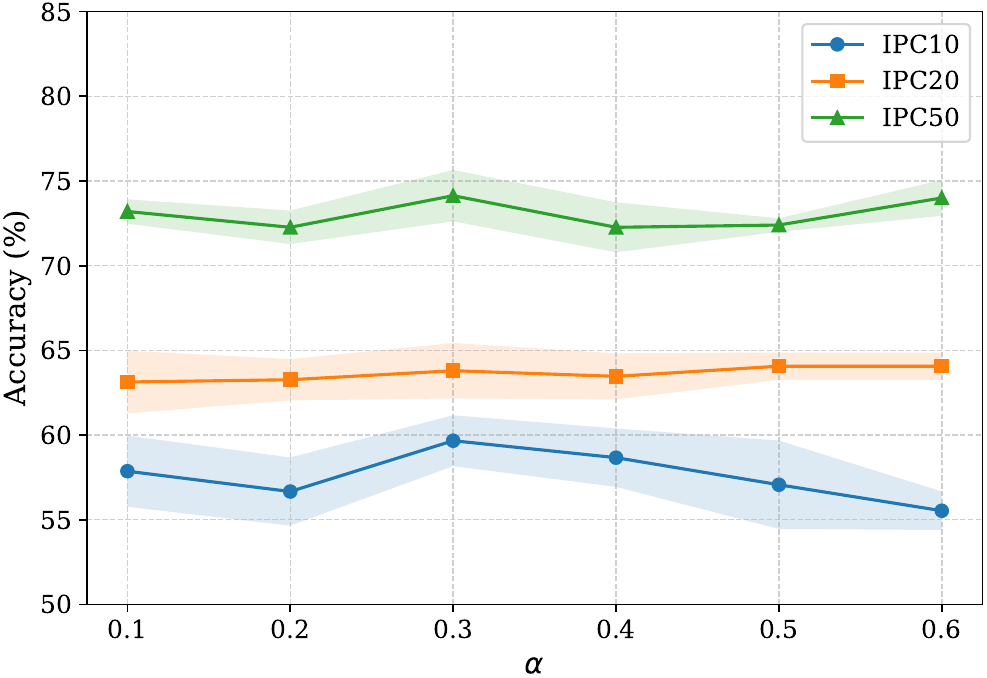}
    \caption{}
    \label{fig:5b}
\end{subfigure}
\hfill
\begin{subfigure}{0.3\linewidth}
    \centering
    \includegraphics[width=\linewidth]{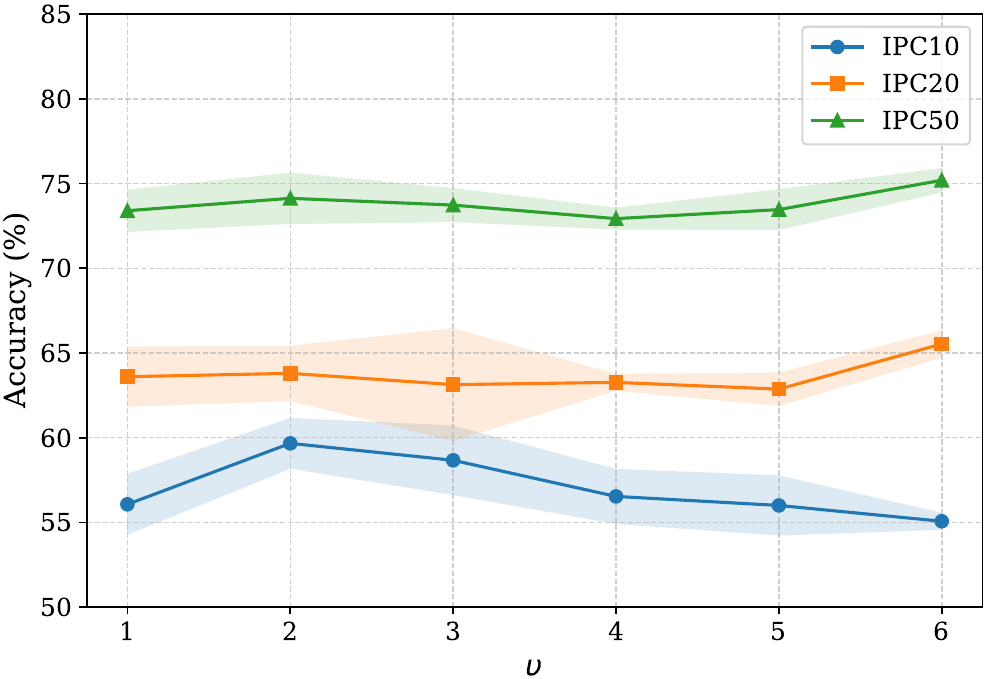}
    \caption{}
    \label{fig:5c}
\end{subfigure}

\caption{Parameter analysis of the saliency-guidance on ImageIDC.}
\label{fig:5}
\end{figure*}

\section{Parameter Analysis}
\label{sec2}
\subsection{Saliency-Guidance Parameters.}
We evaluate the sensitivity of the saliency-guidance parameters $\tau_{\text{cam}}$, $\alpha$, and $\upsilon$ on ImageIDC (Fig.~\ref{fig:5}). For the Grad-CAM++ threshold $\tau_{\text{cam}}$ (Fig.~\ref{fig:5a}), lower IPC settings (10 and 20) are more sensitive to $\tau_{\text{cam}}$, with noticeable fluctuations. A moderate threshold ($\tau_{\text{cam}} = 0.6$) achieves the best balance between noise suppression and object preservation, while IPC = 50 remains stable across a wider range. The parameter $\alpha$ controls the strength of class-discriminative activations (Fig.~\ref{fig:5b}). Increasing $\alpha$ enhances saliency emphasis but may over-concentrate on local regions. IPC = 10 and 50 peaks around $\alpha = 0.3$, while IPC = 20 exhibits comparatively stable behavior. For the voting parameter $\upsilon$ (Fig.~\ref{fig:5c}), low IPC settings exhibit greater sensitivity to threshold variations. IPC = 10 performs best with a small voting threshold at $\upsilon$ = 2, preserving broader salient regions. In contrast, IPC = 20 and 50 benefit from a stricter voting criterion of $\upsilon$ = 6, which filters noisy activations while maintaining structural integrity, achieving peak accuracies of 65.5\%$\pm$0.8 and 75.2\%$\pm$0.7, respectively. Nevertheless, we use $\upsilon=2$ as a shared setting across all IPC
configurations to avoid IPC-specific tuning and preserve performance in
the low-IPC regime. Overall, lower IPC configurations are more sensitive to saliency hyperparameters, whereas higher IPC settings exhibit greater robustness.

\subsection{Prototype Refinement Parameters}
\begin{figure}[t]
    \centering

    \begin{subfigure}{0.4\textwidth}
        \centering
        \includegraphics[width=\linewidth]{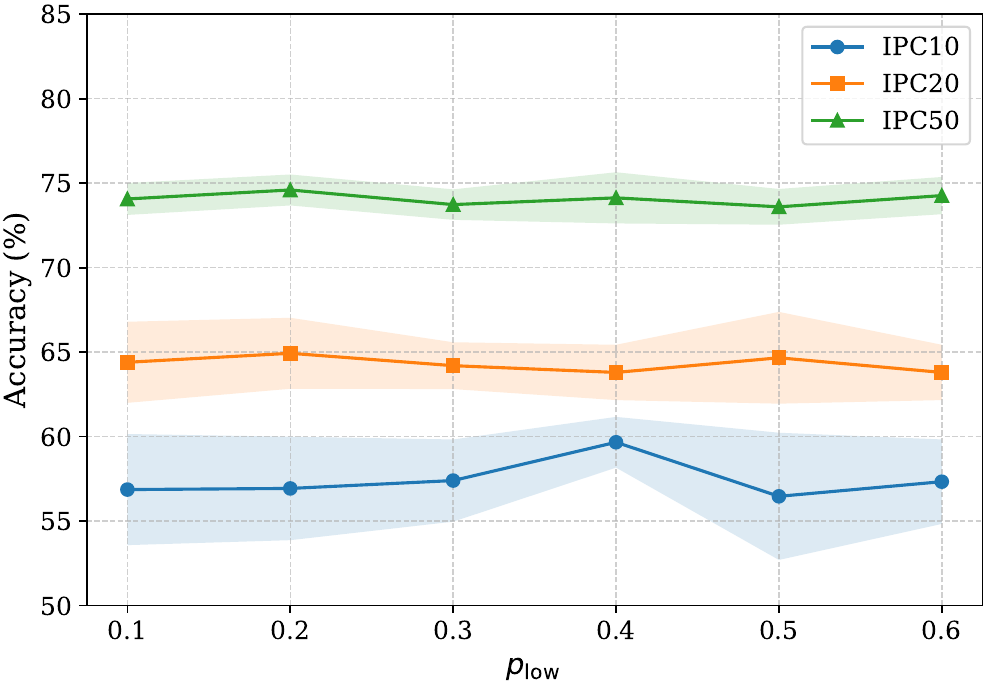}
        \caption{}
        \label{fig:6a}
    \end{subfigure}

    \vspace{2mm}

    \begin{subfigure}{0.4\textwidth}
        \centering
        \includegraphics[width=\linewidth]{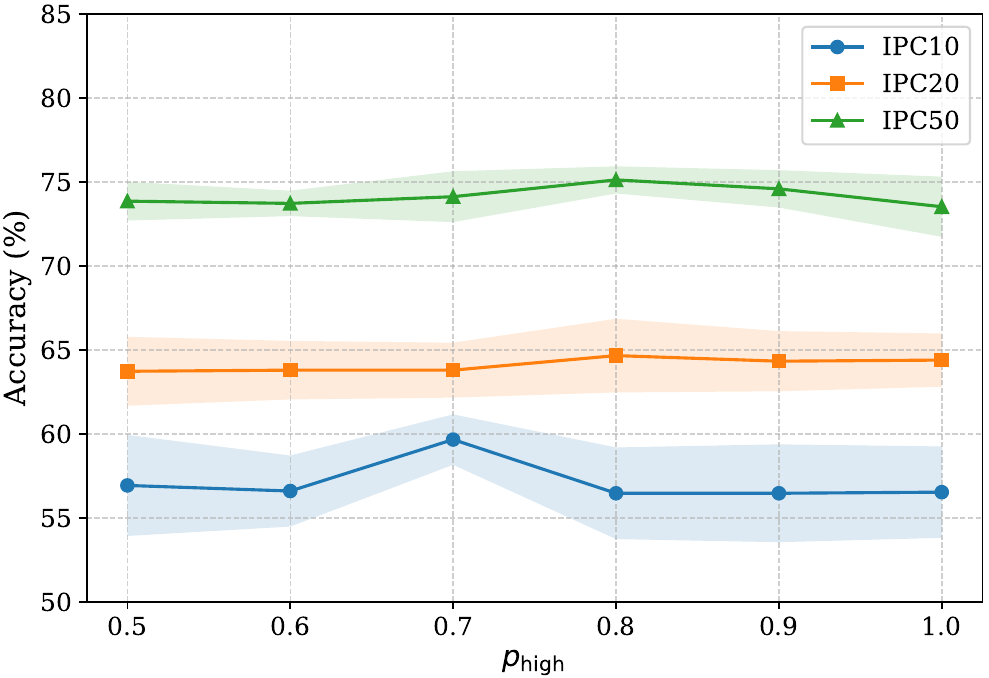}
        \caption{}
        \label{fig:6b}
    \end{subfigure}

    \caption{Parameter analysis of prototype refinement on ImageIDC:
    (a) the lower threshold $p_{\mathrm{low}}$ and
    (b) the upper threshold $p_{\mathrm{high}}$.}
    \label{fig:6}
\end{figure}

We next examine the influence of the prototype refinement confidence thresholds $p_{low}$ and $p_{high}$. These thresholds define an acceptable confidence interval, and prototypes deviating from this interval are iteratively updated until their predicted confidence falls within the prescribed range. As shown in Fig.~\ref{fig:6}\subref{fig:6a}, within the range of $0.1\!-\!0.4$, the performance at IPC$=10$ improves steadily as $p_{low}$ increases. This result suggests that enforcing a minimum confidence level for prototypes can reduce noise and improve prototype quality. However, when $p_{low} \ge 0.5$, the performance declines. This behavior can be attributed to the fact that relying exclusively on highly confident prototypes reduces diversity and weakens sensitivity to boundary cases. In contrast, the results for IPC = 20 and IPC = 50 remain relatively stable across all settings, which indicates that the proposed method is generally robust and is not highly sensitive to this threshold under higher-IPC regimes.
We then investigate the effect of the upper threshold $p_{high}$. As shown in Fig.~\ref{fig:6}\subref{fig:6b}, the performance for IPC = 20 and IPC = 50 gradually improves as $p_{high}$ increases from $0.5$ to $0.8$. In particular, IPC = 50 achieves the best accuracy of 75.1\% at $p_{high}=0.8$, whereas IPC = 20 reaches 64.6\%\,$\pm$\,2.1. These results suggest that using a higher upper threshold can improve performance without sacrificing diversity. By contrast, the curve for IPC = 10 exhibits substantially stronger fluctuations, peaking around $p_{high}=0.7$ and declining at larger values of $p_{high}$. This instability reflects the limited representational capacity in low-IPC regimes, where the student model's performance is highly sensitive to the precise configuration of individual prototypes. In such cases, the delicate balance between boundary coverage and class representativeness makes the distillation process more susceptible to variations in the $p_{high}$ threshold. Based on these observations, we adopt a fixed interval $[p_{low}, p_{high}] = [0.4, 0.7]$ across all IPC settings and datasets, rather than tuning it per configuration. This choice lies close to the empirically favorable region for IPC = 10, where sensitivity to these thresholds is most pronounced, while remaining within the broadly stable range observed for IPC = 20 and IPC = 50. We leave dataset- or IPC-adaptive thresholding as a direction for future work.

\subsection{Analysis of Saliency-Guidance and Hard-Prototype Refinement}
\noindent
We further analyze the joint effect of Saliency Guidance (SG) and
Hard-Prototype (HP) refinement. As shown in
Tab.~\ref{tab:ipc_comparison}, introducing SG leads to moderate
improvements in generative quality and distributional coverage, as
indicated by a slight reduction in FID and increases in Density and
Coverage. These results suggest that SG helps the prototypes better
capture representative regions of the underlying data distribution.

Building on the SG-refined prototypes, the HP module further moves the prototypes toward more challenging
decision-boundary regions. After HP refinement,
FID and Coverage remain broadly comparable, while Precision increases
from $89.20{\pm}1.00$ to $90.13{\pm}1.30$. This suggests that HP
refinement may improve class consistency without substantially altering
the global synthetic distribution. Overall, SG and HP exhibit
complementary effects: SG mainly improves representativeness and
distributional coverage, whereas HP refines prototype difficulty and
discriminability. 
\begin{table}[t]
\centering
{\small
\setlength{\tabcolsep}{0.5pt}
\begin{tabular}{lcccc}
\toprule
Refinement &
Prec. (\%) $\uparrow$ &
Dens. $\uparrow$ &
Cov. (\%) $\uparrow$ &
FID $\downarrow$ \\
\midrule
MGD$^{3}$        & \textbf{90.73{$\pm$0.76}} &  1.17{$\pm$0.10} & 17.90{$\pm$1.30} & 52.37{$\pm$0.89} \\
MGD$^{3}$ + SG & $89.20{\pm1.00}$ & \textbf{1.24{$\pm$0.02}} & $18.39{\pm0.14}$ & \textbf{51.54{$\pm$1.23}}\\
MGD$^{3}$ + SG + HP  & $90.13{\pm1.30}$ & $1.21{\pm0.06}$ & \textbf{18.51{$\pm$1.17}} & {51.57{$\pm$1.34}} \\
\bottomrule
\end{tabular}
}
\vspace{4pt}
\caption{Effects of saliency guidance and hard-prototype
refinement on generation quality. We report Precision,
Density, Coverage, and FID (mean$\pm$std).}
\label{tab:ipc_comparison}
\end{table}
\subsection{Impact of Classifier Ensemble Size}

We further investigate the impact of the classifier ensemble size $M$ on
distillation performance. As shown in Tab.~\ref{tab:config_analysis}, we
compare the original MGD$^{3}$ baseline with our method using
$M\in\{1,2,3,5,10\}$ on ImageWoof under IPC = 10, 20, and 50. Compared
with the single-classifier setting ($M=1$), using multiple classifiers
generally improves performance across different evaluation networks,
demonstrating that classifier ensembles provide complementary
discriminative information and more reliable saliency guidance. However,
the performance does not increase monotonically with $M$, and the optimal
ensemble size varies across IPC settings and architectures.

\begin{table}[t]
\centering

\small
\setlength{\tabcolsep}{1.5pt}
\resizebox{\columnwidth}{!}{
\begin{tabular}{llccc}
\toprule
\textbf{Setting} & \textbf{Network}
& \textbf{IPC10} & \textbf{IPC20} & \textbf{IPC50} \\
\midrule

\multirow{3}{*}{MGD$^{3}$}
& ConvNet-6
& $34.7{\pm}1.1$
& $39.0{\pm}3.5$
& $\mathbf{54.5{\pm}1.6}$ \\

& ResNet-AP-10
& $40.4{\pm}1.9$
& $43.6{\pm}1.6$
& $56.5{\pm}1.9$ \\

& ResNet-18
& $38.5{\pm}2.5$
& $41.9{\pm}2.1$
& $58.3{\pm}1.4$ \\
\midrule

\multirow{3}{*}{$M=1$}
& ConvNet-6
& $34.5{\pm}1.7$
& $38.3{\pm}1.7$
& $51.9{\pm}1.6$ \\

& ResNet-AP-10
& $40.8{\pm}2.7$
& $46.0{\pm}1.1$
& $60.7{\pm}0.9$ \\

& ResNet-18
& $42.9{\pm}2.6$
& $48.5{\pm}1.5$
& $62.8{\pm}0.3$ \\
\midrule

\multirow{3}{*}{$M=2$}
& ConvNet-6
& \underline{$35.3{\pm}0.5$}
& $38.9{\pm}1.0$
& \underline{$53.7{\pm}0.2$} \\

& ResNet-AP-10
& $40.6{\pm}2.8$
& $46.5{\pm}1.8$
& $\mathbf{62.3{\pm}1.4}$ \\

& ResNet-18
& $43.0{\pm}1.3$
& $48.7{\pm}2.0$
& $\mathbf{64.8{\pm}0.7}$ \\
\midrule

\multirow{3}{*}{$M=3$}
& ConvNet-6
& \underline{$35.3{\pm}1.5$}
& $\mathbf{40.1{\pm}1.4}$
& $53.0{\pm}1.1$ \\

& ResNet-AP-10
& $42.0{\pm}2.9$
& $46.9{\pm}0.7$
& $59.7{\pm}1.0$ \\

& ResNet-18
& $43.4{\pm}2.1$
& \underline{$50.4{\pm}1.6$}
& $62.9{\pm}0.2$ \\
\midrule

\multirow{3}{*}{$M=5$}
& ConvNet-6
& $35.1{\pm}1.5$
& $\mathbf{40.1{\pm}1.8}$
& $\mathbf{54.5{\pm}1.0}$ \\

& ResNet-AP-10
& \underline{$42.1{\pm}2.5$}
& \underline{$47.0{\pm}1.1$}
& \underline{$61.5{\pm}1.5$} \\

& ResNet-18
& \underline{$43.5{\pm}3.5$}
& $49.4{\pm}2.0$
& $63.7{\pm}0.8$ \\
\midrule

\multirow{3}{*}{$M=10$}
& ConvNet-6
& $\mathbf{35.9{\pm}2.5}$
& \underline{$39.1{\pm}1.4$}
& \underline{$53.7{\pm}1.3$} \\

& ResNet-AP-10
& $\mathbf{42.4{\pm}2.3}$
& $\mathbf{48.5{\pm}0.5}$
& $61.3{\pm}0.5$ \\

& ResNet-18
& $\mathbf{43.8{\pm}2.9}$
& $\mathbf{50.9{\pm}1.8}$
& \underline{$63.8{\pm}0.7$} \\
\bottomrule
\end{tabular}
}
\vspace{4pt}
\caption{Comparison with MGD$^{3}$ and the effect of ensemble size $M$
under different IPC settings on ImageWoof. The best and second-best
results for each network are highlighted in bold and underlined,
respectively.}
\label{tab:config_analysis}
\end{table}

\begin{figure*}[t] 
\centering 
\includegraphics[width=0.9\textwidth]{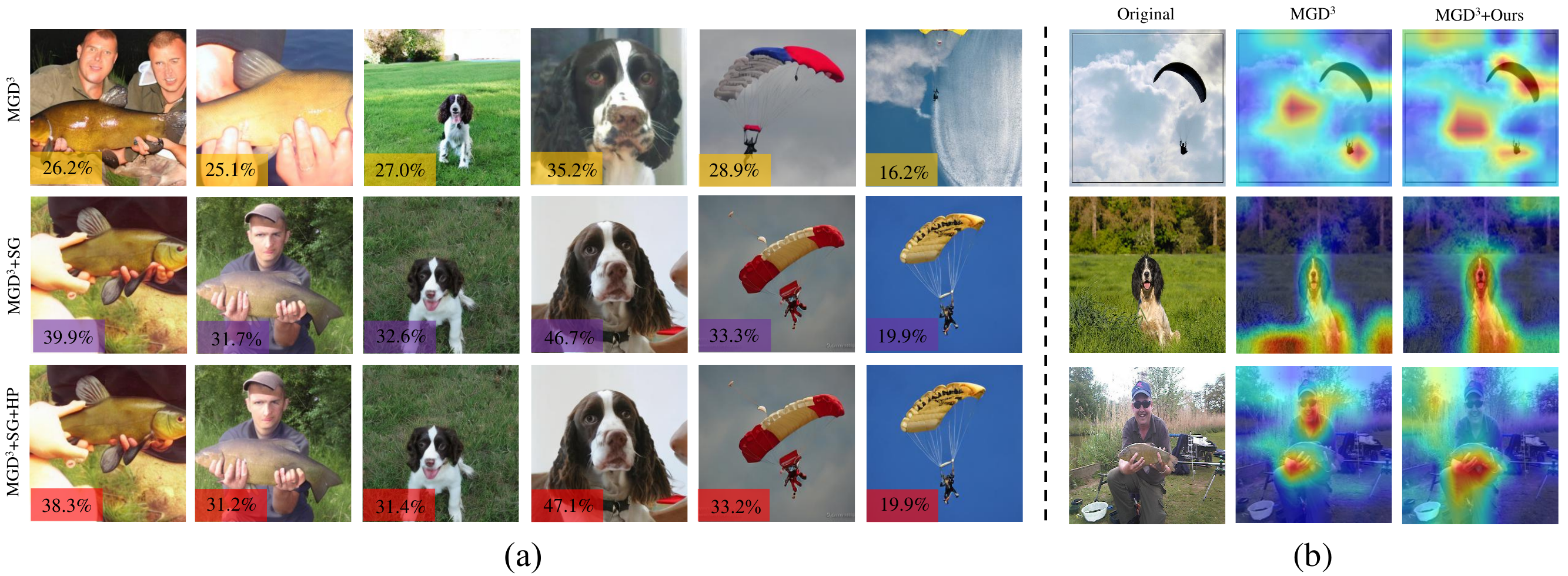} 
\caption{Visualization and saliency analysis (ImageNette, IPC=50). Left: Synthesized samples with the Percentage of Discriminative Areas (PDA) (activation $>0.6$) indicated. Right: Grad-CAM++ on real test data using a model trained on the distilled set. SG and HP denote saliency-guidance and hard-prototype refinement, respectively.
} 
\label{fig3} 
\end{figure*}

\begin{table*}[t]

\centering
\small
\setlength{\tabcolsep}{2pt}
\resizebox{0.85\textwidth}{!}{
\begin{tabular}{l|ccc|ccc}
\toprule
 & \multicolumn{3}{c|}{\textbf{IPC = 10 (0.8\%)}} & \multicolumn{3}{c}{\textbf{IPC = 20 (1.6\%)}} \\
\cmidrule(lr){2-4}\cmidrule(lr){5-7}
 & ConvNet-6 & ResNetAP-10 & ResNet-18 & ConvNet-6 & ResNetAP-10 & ResNet-18 \\
\midrule
Random           & 17.0{$\pm$0.3} & 19.1{$\pm$0.4} & 17.5{$\pm$0.5} & 24.8{$\pm$0.2} & 26.7{$\pm$0.5} & 25.5{$\pm$0.3} \\
Herding         & 17.2{$\pm$0.3} & 19.8{$\pm$0.3} & 16.1{$\pm$0.2} & 24.3{$\pm$0.4} & 27.6{$\pm$0.1} & 24.7{$\pm$0.1} \\
IDC-1           & 24.3{$\pm$0.5} & 25.7{$\pm$0.1} & \underline{25.1{$\pm$0.2}} & 28.8{$\pm$0.3} & 29.9{$\pm$0.2} & 30.2{$\pm$0.2} \\
Minimax         & 22.3{$\pm$0.5} & 24.8{$\pm$0.2} & 22.5{$\pm$0.3} & 29.3{$\pm$0.4} & 32.3{$\pm$0.1} & 31.2{$\pm$0.1} \\
VLCP            & 22.3{$\pm$0.2} & 24.5{$\pm$0.1} & 23.3{$\pm$0.5} & 29.3{$\pm$0.6} & 32.3{$\pm$0.6} & 31.5{$\pm$1.1} \\
D\textsuperscript{4}M
                  & \underline{24.4{$\pm$1.0}} & 26.2{$\pm$0.7} & 24.4{$\pm$1.0} & \textbf{31.7{$\pm$0.3}} & 33.5{$\pm$0.7} & 31.6{$\pm$0.5} \\
D\textsuperscript{4}M+Ours
                  & \textbf{25.0{$\pm$0.4}} & \underline{26.6{$\pm$0.2}} & \textbf{25.5{$\pm$0.3}} & 31.4{$\pm$0.9} & \underline{34.0{$\pm$0.7}} & \underline{33.0{$\pm$1.2}} \\
MGD\textsuperscript{3}
                  & 23.4{$\pm$0.9} & 25.8{$\pm$0.5} & 23.6{$\pm$0.4} & 30.6{$\pm$0.4} & 33.9{$\pm$1.1} & 32.6{$\pm$0.4} \\
MGD\textsuperscript{3}+Ours
                  & 23.9{$\pm$0.4} & \textbf{26.7{$\pm$0.3}} & 24.2{$\pm$0.4} & \underline{31.5{$\pm$0.4}} & \textbf{35.5{$\pm$0.7}} & \textbf{33.5{$\pm$0.9}} \\
\midrule
Full           & 79.9{$\pm$0.4} & 80.3{$\pm$0.2} & 81.8{$\pm$0.7} & 79.9{$\pm$0.4} & 80.3{$\pm$0.2} & 81.8{$\pm$0.7} \\
\bottomrule
\end{tabular}
}
\vspace{4pt}
\caption{Quantitative evaluation on ImageNet-100 against state-of-the-art methods under different IPC settings and model architectures (all at 256 $\times$ 256). The best results are marked as bold; the second best is underlined.}
\label{tab:imgnet100}
\end{table*}

Although $M=10$ achieves the best result in several settings, its gain
over $M=5$ is limited while requiring twice as many classifiers. In
particular, on ConvNet-6, $M=5$ achieves $40.1{\pm}1.8\%$ under IPC = 20,
outperforming MGD$^{3}$ at $39.0{\pm}3.5\%$, and obtains
$54.5{\pm}1.0\%$ under IPC = 50, matching its mean accuracy while showing
lower variance than MGD$^{3}$ at $54.5{\pm}1.6\%$. Moreover, $M=5$
maintains competitive performance across the other evaluation networks.
Therefore, considering both effectiveness and computational efficiency,
we use $M=5$ for the experiments on ImageWoof.

\section{Additional Evaluation Results}
\label{sec3}
\noindent\textbf{ImageNet-100.}
We further evaluate our approach on ImageNet-100 under multiple IPC
settings and evaluation architectures. As shown in
Tab.~\ref{tab:imgnet100}, incorporating the proposed saliency-guided
prototype module generally improves the mean accuracy of both
D$^{4}$M and MGD$^{3}$. The gains reach up to 1.4\% when integrated
with D$^{4}$M and up to 1.6\% when integrated with MGD$^{3}$.
Overall, our module improves the mean accuracy in 11 of the 12
evaluated settings and achieves the best or second-best result in nine
settings, suggesting broad compatibility across different diffusion-based
baselines, evaluation architectures, and IPC configurations.

\noindent\textbf{CIFAR-10 and CIFAR-100.}
We further evaluate our method on the low-resolution CIFAR-10 and CIFAR-100 benchmarks. As shown in Tab.~\ref{tab3}, integrating our saliency-guided prototype module into D$^{4}$M consistently improves performance across all IPC settings. On CIFAR-10, our method improves upon D$^{4}$M by 4.5\% and 2.5\% at IPC = 10 and 50, respectively. On CIFAR-100, it achieves gains of 2.0\% and 1.5\% at the corresponding IPC settings. These consistent improvements demonstrate the effectiveness and generalizability of our module on low-resolution datasets with different numbers of classes.

\noindent\textbf{Visualizations.}
Visual comparisons of synthesized samples and Grad-CAM++ maps are provided
in Fig.~\ref{fig3}. The integration of saliency guidance (SG) increases
the Percentage of Discriminative Areas (PDA) compared with the
MGD$^{3}$ baseline, providing a more informative basis for downstream
feature learning. The slight decrease in PDA after hard-prototype (HP)
refinement suggests that HP does not simply maximize discriminative-area
coverage. Instead, it adjusts overly easy prototypes toward a more
challenging yet class-consistent regime, reducing redundant
high-confidence patterns while preserving informative semantic content.

Qualitative Grad-CAM++ visualizations in Fig.~\ref{fig3}(b) further support
this observation. The model trained on the distilled data generated by
our full framework (MGD$^{3}$ + Ours) exhibits more concentrated and
semantically aligned attention maps on real test data than the baseline.
These results suggest that SG enhances class-discriminative regions,
while HP encourages the model to learn more challenging and informative
features. Together, the two components reduce background interference
and improve the learning of class-relevant representations.

\noindent\textbf{Computational Cost.}
Tab.~\ref{tab:runtime_breakdown} provides a detailed comparison of the computational cost of different diffusion-based distillation methods. Unlike Minimax and MGD\textsuperscript{3} on ImageWoof, our method does not fine-tune the diffusion backbone. Its main additional cost comes from training lightweight classifiers on VAE latents, whereas saliency extraction, hard-prototype refinement, clustering, and image generation introduce relatively limited overhead.

\begin{table}[t]

\centering
\small
\begin{tabularx}{0.48\textwidth}{
  >{\centering\arraybackslash}c
  >{\centering\arraybackslash}c
  >{\centering\arraybackslash}X
  >{\centering\arraybackslash}X
  >{\centering\arraybackslash}X
  >{\centering\arraybackslash}X}
\toprule
Dataset & IPC & SRe\textsuperscript{2}L & RDED & $\text{D}^4\text{M}$  & Ours \\
\midrule
\multirow{2}{*}{CIFAR-10}
  & 10 & 29.3{±0.5} & 37.1{±0.3}  & 36.5{±1.3} & \textbf{41.0{±0.4}} \\
  & 50 & 45.0{±0.7} & 62.1{±0.1} & \underline{66.6{±1.0}} & \textbf{69.1{±0.3} }\\
\midrule
\multirow{2}{*}{CIFAR-100}
  & 10 & 27.0{±0.4} & 42.6{±0.2}  & \underline{52.5{±0.7}} & \textbf{54.5{±0.6} }\\
  & 50 & 50.2{±0.4} & 62.6{±0.1}  & \underline{66.4{±0.1}} & \textbf{67.9{±0.1}} \\
\bottomrule
\end{tabularx}
\vspace{4pt}
\caption{Quantitative evaluation on CIFAR-10 and CIFAR-100.}
\label{tab3}
\end{table}

\begin{table}[b]
\centering

{\small
\setlength{\tabcolsep}{0.8pt}

\begin{tabular}{@{}lccccc@{}}
\toprule
\textbf{Component} 
& {Minimax} 
& {MGD$^3$} 
& {MGD$^3$} 
& {Ours} 
& {Ours} \\
&  
&  
& {(Woof)} 
& {($M=10$)} 
& {($M=5$)} \\
\midrule
Classifier training
& - & - & - & $\sim$30 & $\sim$15 \\
DiT fine-tuning
& $\sim$30 & - & $\sim$30 & - & - \\
Grad-CAM++/masking
& - & - & - & $\sim$0.5 & $\sim$0.5 \\
HP refinement
& - & - & - & $\sim$1 & $\sim$1 \\
Clustering/inference 
& $\sim$4 & $\sim$5 & $\sim$5 & $\sim$5.5 & $\sim$5.5 \\
\midrule
\textbf{Total cost}
& $\sim$34 & $\sim$5 & $\sim$35 & $\sim$37 & $\sim$22 \\
\bottomrule
\end{tabular}
}
\vspace{4pt}
\caption{Runtime breakdown of different components. All runtime values are reported in minutes.}
\label{tab:runtime_breakdown}
\end{table}

With an ensemble size of \(M=10\), the complete pipeline requires approximately 37 minutes, corresponding to an amortized end-to-end cost of \(4.44\,\mathrm{s/image}\) when generating 500 images. For ImageWoof, we adopt \(M=5\), which reduces the total runtime to approximately 22 minutes, corresponding to \(2.64\,\mathrm{s/image}\), while maintaining competitive performance, as shown in Tab.~\ref{tab:config_analysis}. Therefore, the \(4.44\,\mathrm{s/image}\) reported in Sec.~5.3 of the main paper refers to the general \(M=10\) configuration, whereas the actual ImageWoof configuration incurs an amortized cost of \(2.64\,\mathrm{s/image}\). Compared with MGD\textsuperscript{3} on ImageWoof, the adopted $M=5$ configuration reduces the total runtime from approximately 35 minutes to 22 minutes. The computational resources are also allocated differently: MGD\textsuperscript{3} spends most of its additional cost on DiT fine-tuning, whereas our method only trains lightweight classifiers and keeps the generative backbone frozen. Moreover, once trained on a given dataset, the classifier ensemble can be reused for prototype construction in different generative distillation pipelines, such as D\textsuperscript{4}M and MGD\textsuperscript{3}. These results show that the ensemble size offers a flexible trade-off between computational cost and distillation performance.

\begin{figure*}[t]
\centering
\includegraphics[width=0.9\textwidth]{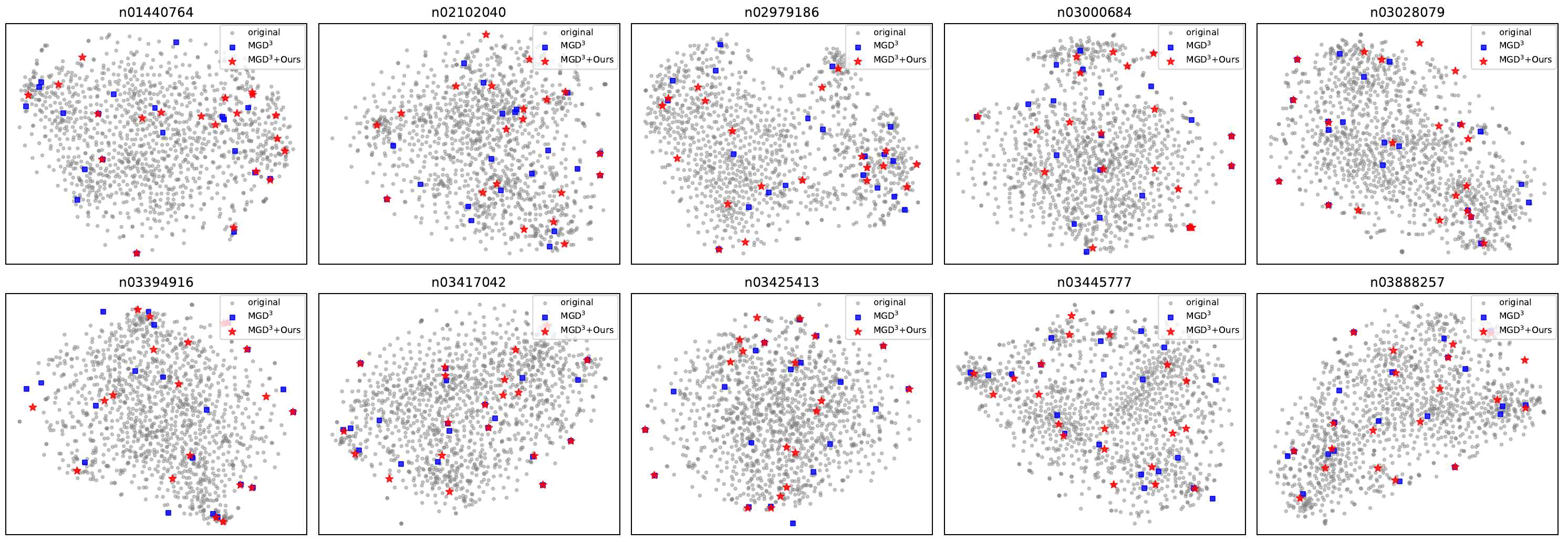}
\vspace{4pt}
\tikz{\draw[dashed, line width=0.5pt] (0,0) -- (0.9\textwidth,0);}
\vspace{4pt}
\includegraphics[width=0.9\textwidth]{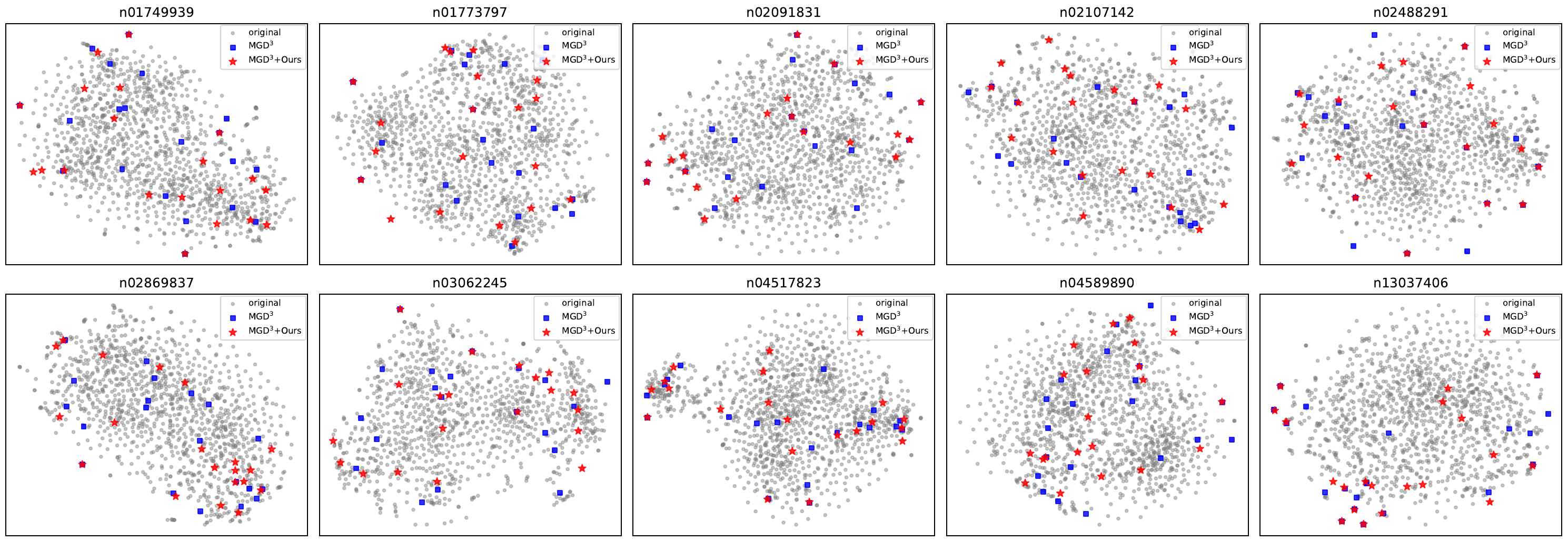}
\caption{t-SNE visualizations of the distilled datasets on ImageNette (top) and ImageIDC (bottom).}
\label{fig:tsne}
\end{figure*}

\section{Visualization via t-SNE}
\label{sec4}
To examine the coverage and representativeness of the distilled datasets, we visualize t-SNE plots of the distilled samples on ImageNette and ImageIDC under IPC = 20. As shown in Fig.~\ref{fig:tsne}, the ImageNette class n03888257 (second row, fifth column) and the ImageIDC class n02107142 (third row, fourth column) clearly illustrate the distributional differences between the two distilled datasets produced by MGD$^{3}$ and our method. In both cases, many samples distilled by MGD$^{3}$ (blue) cluster tightly in a small region of the latent space, leaving parts of the real-data distribution under-represented. In contrast, the distilled datasets generated by our method (red) are more broadly distributed across the feature space and align more closely with high-density regions of the real data. This indicates that our approach yields distilled datasets whose empirical distributions more faithfully approximate those of the original data, thereby improving representativeness.

\section{Visualization of Distilled Datasets}
\label{sec5}
\begin{figure*}[h]
    \centering

    \setlength{\tabcolsep}{8pt}

    \begin{tabular}{c:c}

        \begin{minipage}[t]{0.43\textwidth}
            \centering

            \textbf{ImageNette}\\[3pt]

            \setlength{\tabcolsep}{1pt}
            \begin{tabular}{@{}cccc@{}}

                \includegraphics[width=0.24\linewidth]
                {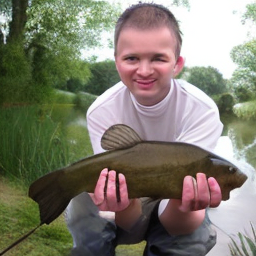}
                &
                \includegraphics[width=0.24\linewidth]
                {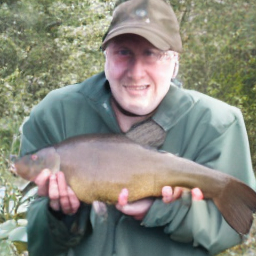}
                &
                \includegraphics[width=0.24\linewidth]
                {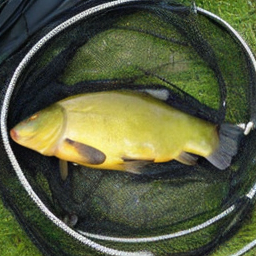}
                &
                \includegraphics[width=0.24\linewidth]
                {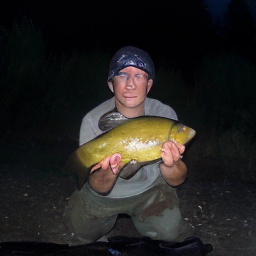}
                \\

                \includegraphics[width=0.24\linewidth]
                {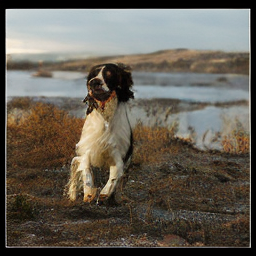}
                &
                \includegraphics[width=0.24\linewidth]
                {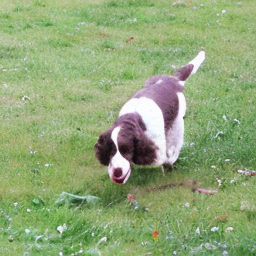}
                &
                \includegraphics[width=0.24\linewidth]
                {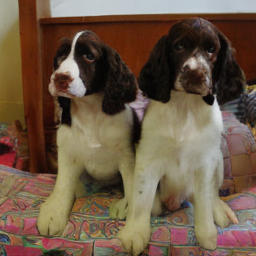}
                &
                \includegraphics[width=0.24\linewidth]
                {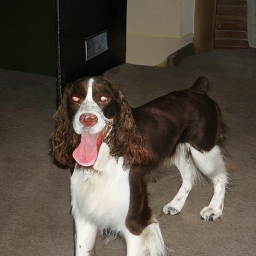}
                \\

                \includegraphics[width=0.24\linewidth]
                {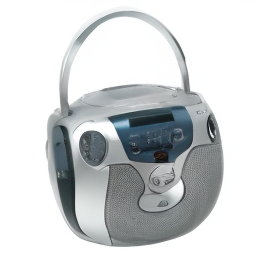}
                &
                \includegraphics[width=0.24\linewidth]
                {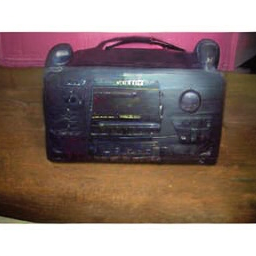}
                &
                \includegraphics[width=0.24\linewidth]
                {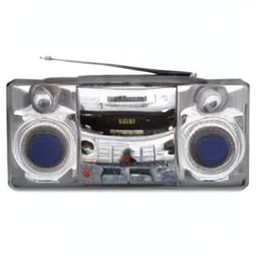}
                &
                \includegraphics[width=0.24\linewidth]
                {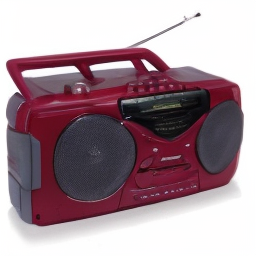}
                \\

                \includegraphics[width=0.24\linewidth]
                {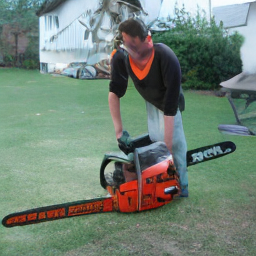}
                &
                \includegraphics[width=0.24\linewidth]
                {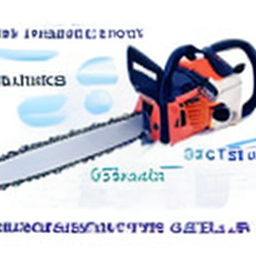}
                &
                \includegraphics[width=0.24\linewidth]
                {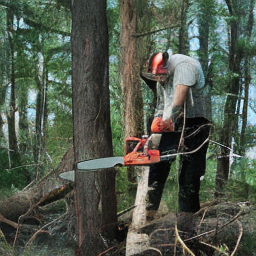}
                &
                \includegraphics[width=0.24\linewidth]
                {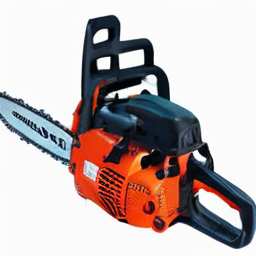}
                \\

                \includegraphics[width=0.24\linewidth]
                {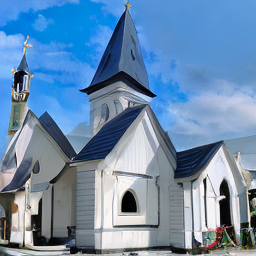}
                &
                \includegraphics[width=0.24\linewidth]
                {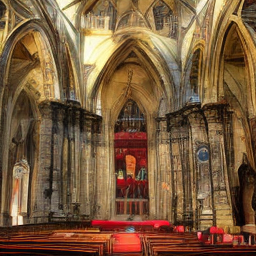}
                &
                \includegraphics[width=0.24\linewidth]
                {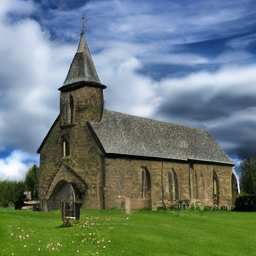}
                &
                \includegraphics[width=0.24\linewidth]
                {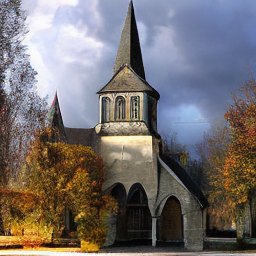}
                \\

                \includegraphics[width=0.24\linewidth]
                {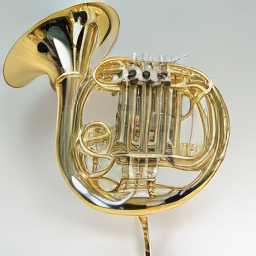}
                &
                \includegraphics[width=0.24\linewidth]
                {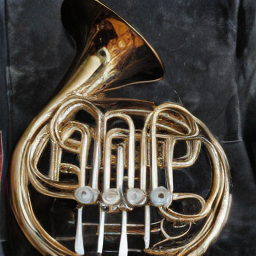}
                &
                \includegraphics[width=0.24\linewidth]
                {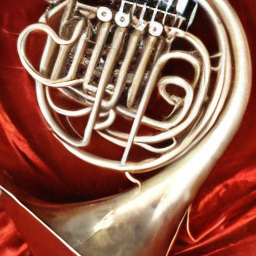}
                &
                \includegraphics[width=0.24\linewidth]
                {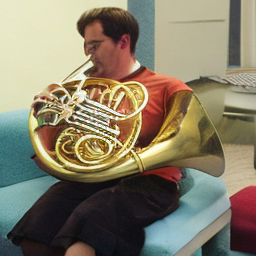}
                \\

                \includegraphics[width=0.24\linewidth]
                {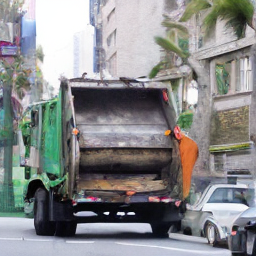}
                &
                \includegraphics[width=0.24\linewidth]
                {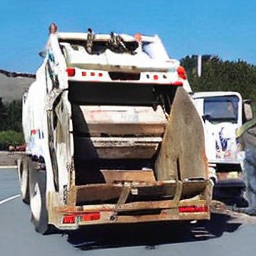}
                &
                \includegraphics[width=0.24\linewidth]
                {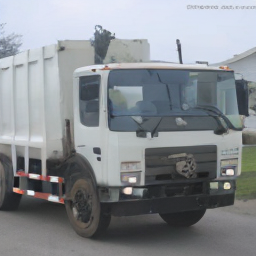}
                &
                \includegraphics[width=0.24\linewidth]
                {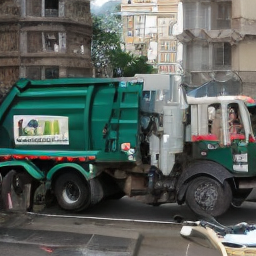}
                \\

                \includegraphics[width=0.24\linewidth]
                {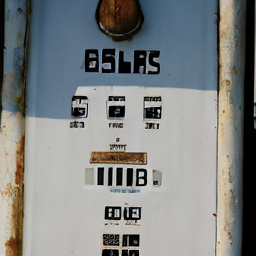}
                &
                \includegraphics[width=0.24\linewidth]
                {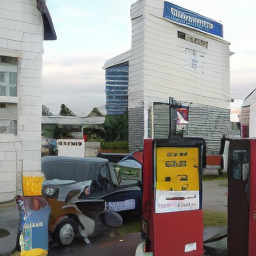}
                &
                \includegraphics[width=0.24\linewidth]
                {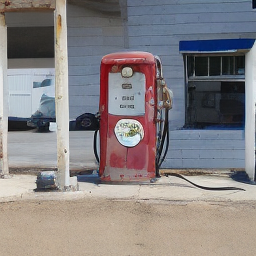}
                &
                \includegraphics[width=0.24\linewidth]
                {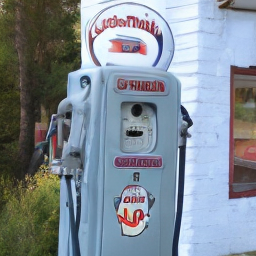}
                \\

                \includegraphics[width=0.24\linewidth]
                {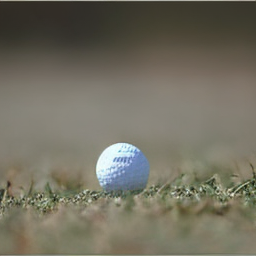}
                &
                \includegraphics[width=0.24\linewidth]
                {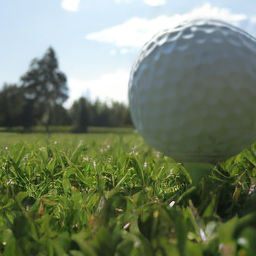}
                &
                \includegraphics[width=0.24\linewidth]
                {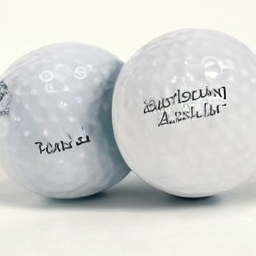}
                &
                \includegraphics[width=0.24\linewidth]
                {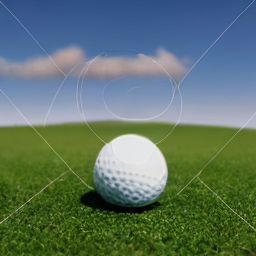}
                \\

                \includegraphics[width=0.24\linewidth]
                {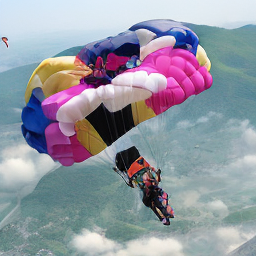}
                &
                \includegraphics[width=0.24\linewidth]
                {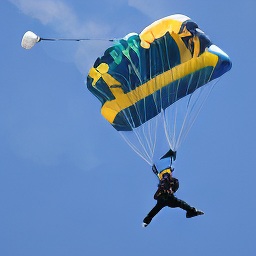}
                &
                \includegraphics[width=0.24\linewidth]
                {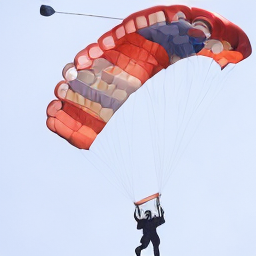}
                &
                \includegraphics[width=0.24\linewidth]
                {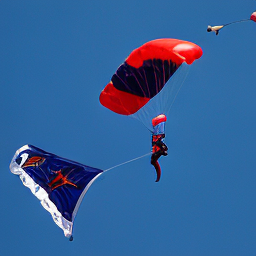}
                \\

            \end{tabular}
        \end{minipage}

        &

        \begin{minipage}[t]{0.43\textwidth}
            \centering

            \textbf{ImageIDC}\\[3pt]

            \setlength{\tabcolsep}{1pt}
            \begin{tabular}{@{}cccc@{}}

                \includegraphics[width=0.24\linewidth]
                {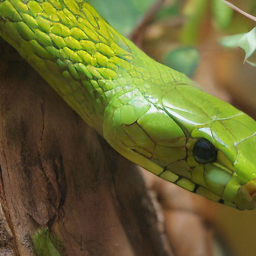}
                &
                \includegraphics[width=0.24\linewidth]
                {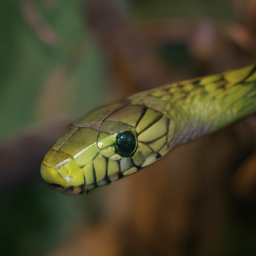}
                &
                \includegraphics[width=0.24\linewidth]
                {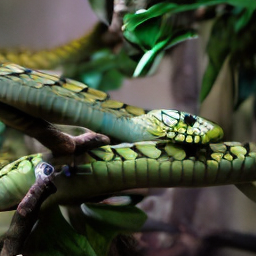}
                &
                \includegraphics[width=0.24\linewidth]
                {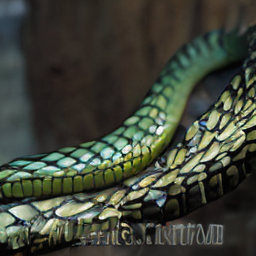}
                \\

                \includegraphics[width=0.24\linewidth]
                {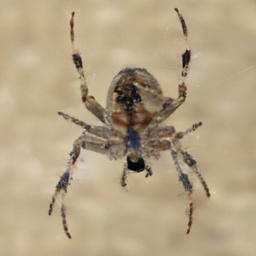}
                &
                \includegraphics[width=0.24\linewidth]
                {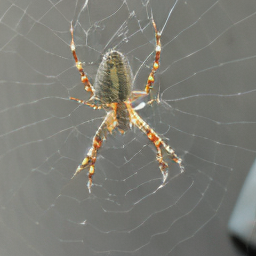}
                &
                \includegraphics[width=0.24\linewidth]
                {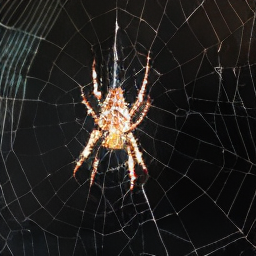}
                &
                \includegraphics[width=0.24\linewidth]
                {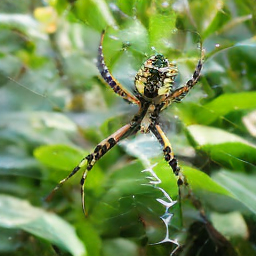}
                \\

                \includegraphics[width=0.24\linewidth]
                {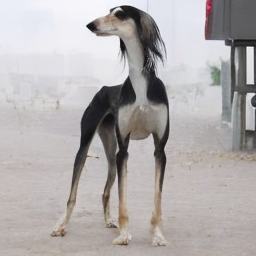}
                &
                \includegraphics[width=0.24\linewidth]
                {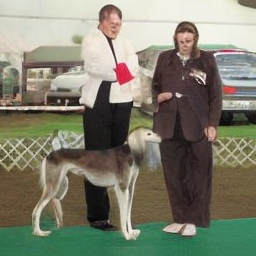}
                &
                \includegraphics[width=0.24\linewidth]
                {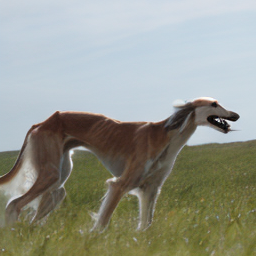}
                &
                \includegraphics[width=0.24\linewidth]
                {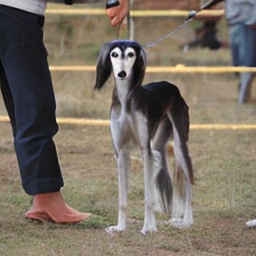}
                \\

                \includegraphics[width=0.24\linewidth]
                {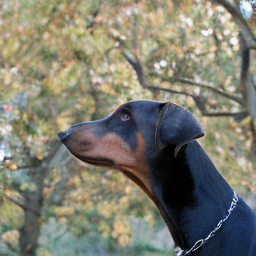}
                &
                \includegraphics[width=0.24\linewidth]
                {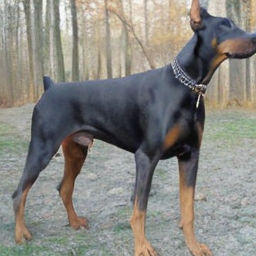}
                &
                \includegraphics[width=0.24\linewidth]
                {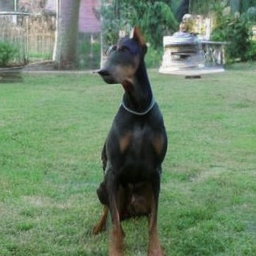}
                &
                \includegraphics[width=0.24\linewidth]
                {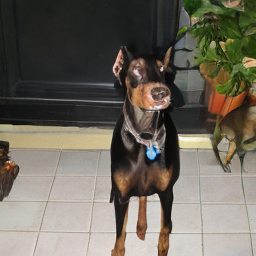}
                \\

                \includegraphics[width=0.24\linewidth]
                {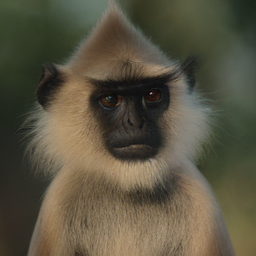}
                &
                \includegraphics[width=0.24\linewidth]
                {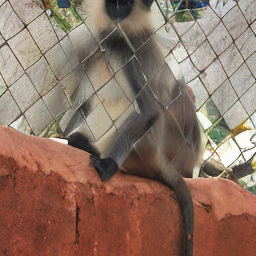}
                &
                \includegraphics[width=0.24\linewidth]
                {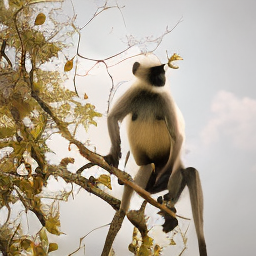}
                &
                \includegraphics[width=0.24\linewidth]
                {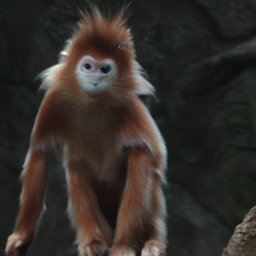}
                \\

                \includegraphics[width=0.24\linewidth]
                {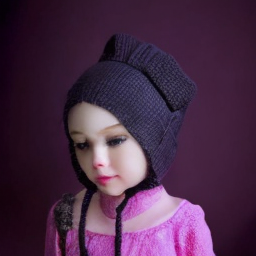}
                &
                \includegraphics[width=0.24\linewidth]
                {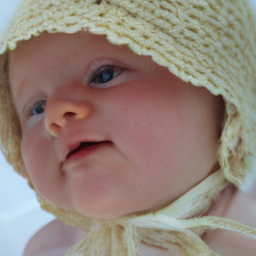}
                &
                \includegraphics[width=0.24\linewidth]
                {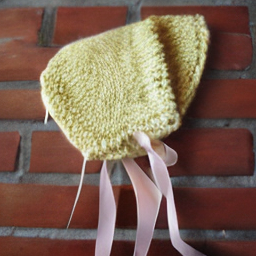}
                &
                \includegraphics[width=0.24\linewidth]
                {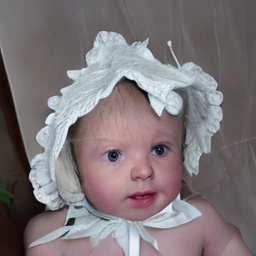}
                \\

                \includegraphics[width=0.24\linewidth]
                {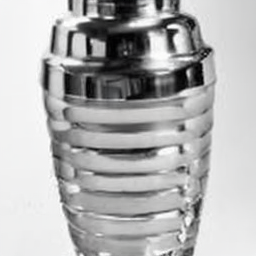}
                &
                \includegraphics[width=0.24\linewidth]
                {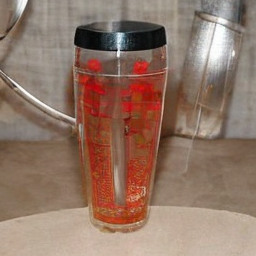}
                &
                \includegraphics[width=0.24\linewidth]
                {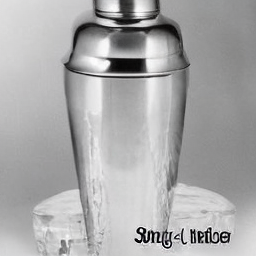}
                &
                \includegraphics[width=0.24\linewidth]
                {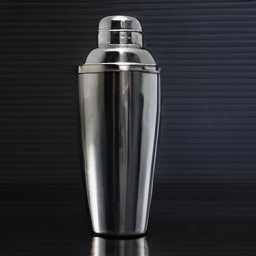}
                \\

                \includegraphics[width=0.24\linewidth]
                {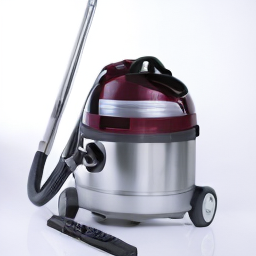}
                &
                \includegraphics[width=0.24\linewidth]
                {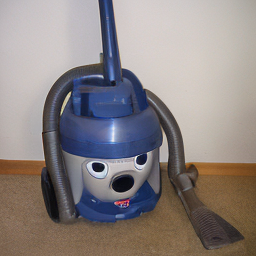}
                &
                \includegraphics[width=0.24\linewidth]
                {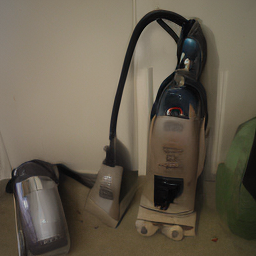}
                &
                \includegraphics[width=0.24\linewidth]
                {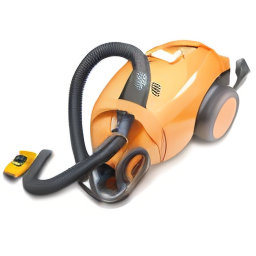}
                \\

                \includegraphics[width=0.24\linewidth]
                {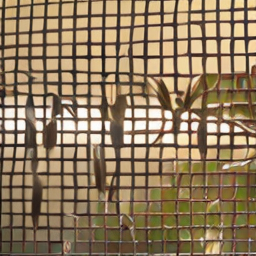}
                &
                \includegraphics[width=0.24\linewidth]
                {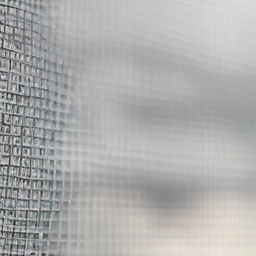}
                &
                \includegraphics[width=0.24\linewidth]
                {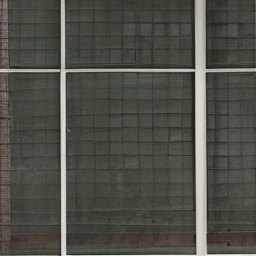}
                &
                \includegraphics[width=0.24\linewidth]
                {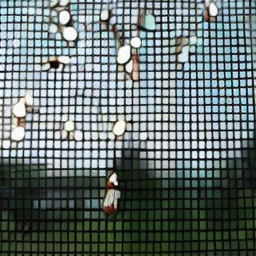}
                \\

                \includegraphics[width=0.24\linewidth]
                {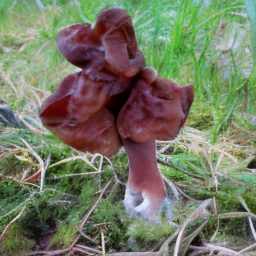}
                &
                \includegraphics[width=0.24\linewidth]
                {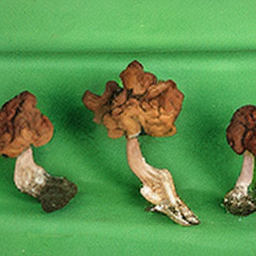}
                &
                \includegraphics[width=0.24\linewidth]
                {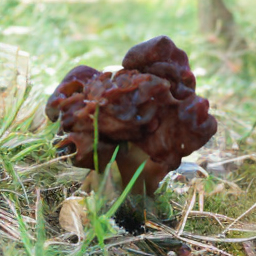}
                &
                \includegraphics[width=0.24\linewidth]
                {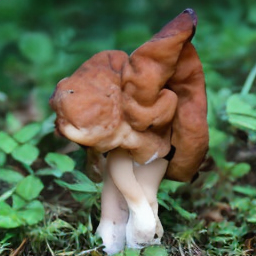}
                \\

            \end{tabular}
        \end{minipage}

    \end{tabular}

    \caption{Visualization of synthetic images at
    $256 \times 256$ resolution and IPC = 10.
    The left and right panels show samples distilled from
    ImageNette and ImageIDC, respectively.}
    \label{fig:additional_synthetic_images}
\end{figure*}

Beyond quantitative evaluation, we also provide qualitative insight by visualizing additional distilled samples without manual selection. Samples from ImageNette and ImageIDC at IPC = 10 are shown in Fig.~\ref{fig:additional_synthetic_images}. Across different datasets, the distilled images preserve class-discriminative details consistent with the target categories and exhibit reasonable intra-class variation, rather than collapsing to a small set of prototypical appearances. Moreover, the overall visual quality is comparable to that of real images, with few noticeable artifacts. These observations further support the conclusion that our method produces visually plausible and diverse synthetic datasets for downstream training.

\end{document}